\documentclass[lettersize,journal]{IEEEtran}
\usepackage{amssymb}
\usepackage{multirow}
\usepackage{booktabs}
\usepackage{enumerate}
\usepackage{setspace} 
\usepackage[ruled,linesnumbered]{algorithm2e}

\usepackage{amsmath,amsfonts}
\usepackage{array}
\usepackage{textcomp}
\usepackage{stfloats}
\usepackage{url}
\usepackage{verbatim}
\usepackage{graphicx}
\usepackage{cite}
\usepackage[colorlinks, citecolor=black]{hyperref}

\usepackage[capitalize]{cleveref}
\crefname{section}{Sec.}{Secs.}
\Crefname{section}{Section}{Sections}
\Crefname{table}{Table}{Tables}

\usepackage{ mathrsfs }
\usepackage{multirow,diagbox}
\usepackage{verbatim}
\usepackage{color}
\usepackage{colortbl}
\usepackage{makecell}
\usepackage{cellspace}
\usepackage{stfloats}
\usepackage{longtable}

\usepackage{array,multirow,graphicx}
\usepackage{color}
\usepackage{float}
\usepackage{makecell}
\usepackage[x11names,dvipsnames,table]{xcolor} 
\usepackage{colortbl}
\usepackage{multirow}
\usepackage[font=footnotesize]{caption}
\usepackage{subcaption}
\usepackage{lipsum}
\definecolor{mygray}{gray}{0.6}
\definecolor{myblue}{rgb}{0.8,0.85,1}
\usepackage{array}
\newcolumntype{L}[1]{>{\raggedright\let\newline\\\arraybackslash\hspace{0pt}}m{#1}}
\newcolumntype{C}[1]{>{\centering\let\newline\\\arraybackslash\hspace{0pt}}m{#1}}
\newcolumntype{R}[1]{>{\raggedleft\let\newline\\\arraybackslash\hspace{0pt}}m{#1}}

\usepackage{array, tabularx, boldline}
\usepackage{eurosym}
\usepackage{amstext} 
\DeclareRobustCommand{\officialeuro}{%
  \ifmmode\expandafter\text\fi
  {\fontencoding{U}\fontfamily{eurosym}\selectfont e}}

\begin{document}
\title{\huge AST: Effective Dataset Distillation through Alignment with Smooth and High-Quality Expert Trajectories}

\author{Jiyuan Shen, Wenzhuo Yang, and Kwok-Yan Lam\\
Nanyang Technological University, Singapore\\
{\tt\small jiyuan001@e.ntu.edu.sg, \{wenzhuo.yang, kwokyan.lam\}@ntu.edu.sg}
%
%
}



\maketitle
\IEEEpubidadjcol

\begin{abstract}

Training large AI models typically requires large-scale datasets in the machine learning process, making training and parameter-tuning process both time-consuming and costly. Some researchers address this problem by carefully synthesizing a very small number of highly representative and informative samples from real-world datasets. This approach, known as Dataset Distillation (DD), proposes a perspective for data-efficient learning. Despite recent progress in this field, the performance of existing methods still cannot meet expectations, and distilled datasets cannot effectively replace original datasets. In this paper, unlike previous methods that focus solely on improving the effectiveness of student distillation, we recognize and leverage the important mutual influence between expert and student models. We observed that the smoothness of expert trajectories has a significant impact on subsequent student parameter alignment. Based on this, we propose an effective DD framework named AST, standing for \underline{\textit{A}}lignment with \underline{\textit{S}}mooth and high-quality expert \underline{\textit{T}}rajectories. We devise the integration of clipping loss and gradient penalty to regulate the rate of parameter changes in expert trajectory generation. To further refine the student parameter alignment with expert trajectory, we put forward representative initialization for the synthetic dataset and balanced inner-loop loss in response to the sensitivity exhibited towards randomly initialized variables during distillation. We also propose two enhancement strategies, namely intermediate matching loss and weight perturbation, to mitigate the potential occurrence of cumulative errors. We conduct extensive experiments on datasets of different scales, sizes, and resolutions. The results demonstrate that the proposed method significantly outperforms prior methods.

\end{abstract}


\section{Introduction}
\label{intro}


Nowadays, large machine learning models have demonstrated unprecedented results in many fields. 
Behind the success of large models, large-scale datasets become an indispensable driving force.
Despite the great potential of such approaches, we still wonder about the underlying principles and limitations. Do datasets have to be big? Can the information from a large dataset be distilled and integrated? Or, can training on ``small data'' be equally successful?

As a result, some researchers explore data-efficient methods, including coreset selection \cite{sener2017active,aljundi2019gradient,agarwal2004approximating}, dataset pruning \cite{cohn1996active,felzenszwalb2010object,lapedriza2013all}, and instance selection \cite{olvera2010review}. 
They summarize the entire dataset by only selecting the ``valuable'' data for model training. Yet, these methods rely on greedy algorithms guided by heuristics \cite{chen2012super,aljundi2019gradient,sener2017active,Forgetting}, resulting in significant performance drops, 
and the selection process is not a way of unearthing and integrating high-level information.


\begin{figure}
    \centering
    \includegraphics[width=\linewidth]{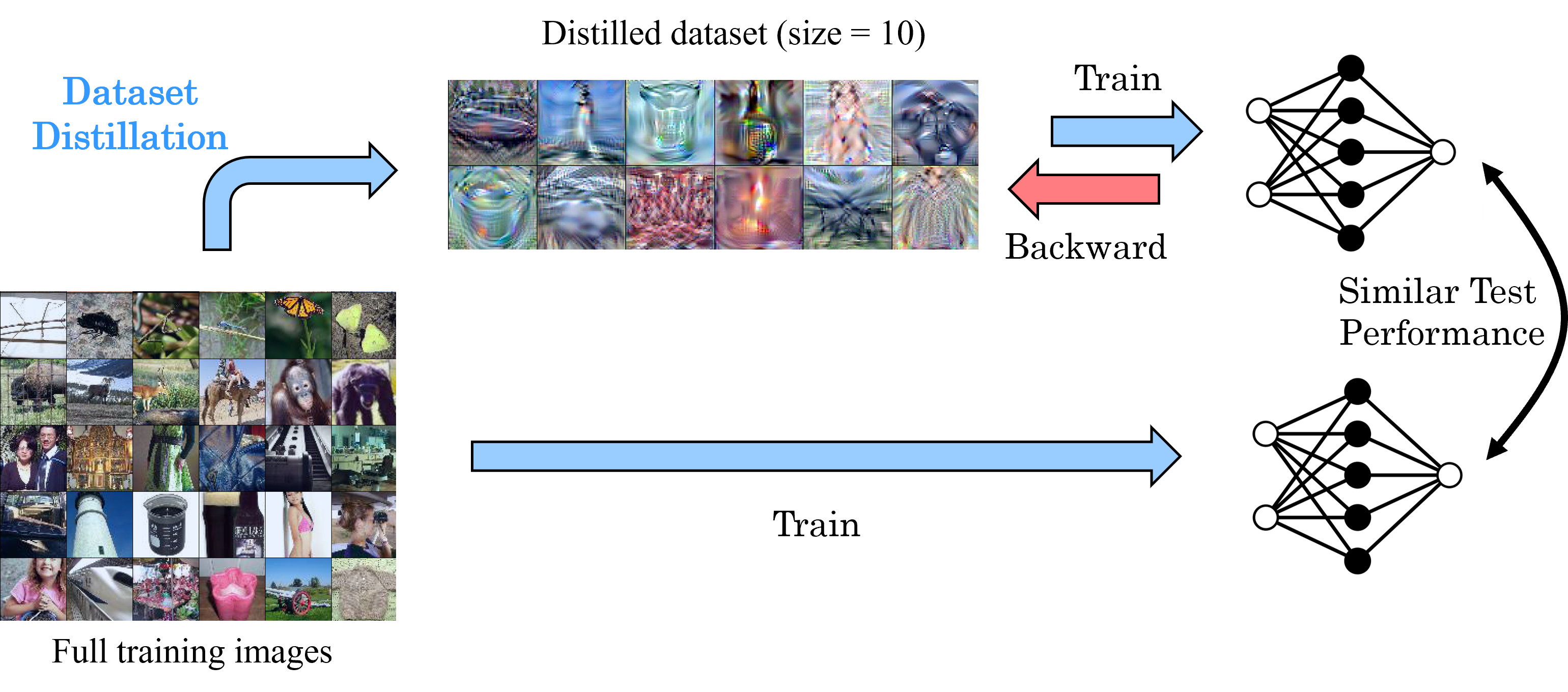}
    \caption{Dataset distillation aims to generate a compact synthetic dataset that allows a model trained on it to achieve a comparable test performance to a model trained on the entire real training set.}
    \label{fig:intro}
\end{figure}

Recently, a new data-efficient method called Dataset Distillation (DD) has emerged as a competitive alternative with promising results \cite{chen2023comprehensive,sachdeva2023data,cui2022dc,yu2023dataset}. DD uses a learning method to distill a large training set (expert) into a small synthetic set (student), upon which machine learning models are trained from scratch, and similar testing performance on the validation set is expected to be preserved, as shown in \cref{fig:intro}. DD has aroused significant interest from the machine learning community with various downstream applications, such as federated learning~\cite{pi2022dynafed,liu2023meta,zhou2020distilled}, neural architecture search~\cite{medvedev2021tabular,such2020generative}, and privacy-preserving tasks~\cite{loo2023attack,liu2023backdoor,chen2022privacy}, and etc. 

As one of the state-of-the-art frameworks for dataset distillation, Matching Training Trajectories (MTT)~\cite{mtt} distinguishes itself by long-range matching characteristics. It employs training trajectories from large training sets as experts, which is an efficient method of information extraction compared to using pixel-level data. Therefore, MTT encompasses two primary phases: expert trajectory generation (buffer) and student parameter alignment (distillation). 

Nevertheless, despite recent advancements, the research works~\cite{mtt,du2022minimizing,tesla} continue to utilize simplistic optimizer for generating weak expert trajectories. Moreover, the alignment process suffers from insufficient supervision, as solely relying on the final weight matching loss is suboptimal for the long-range matching process. Furthermore, nearly all DD algorithms \cite{wang2018dataset,GM,DM} fail to adequately consider the compatibility of experts for students. This oversight neglects potential relationships that could consistently enhance the performance of both students and experts.

To tackle the catastrophic discrepancy arising from a more competent expert and to optimize the subsequent distillation process further, thereby making DD more efficient and effective, we identify three primary reasons that contribute to the reduced performance of the distilled dataset. First, we observe that the smoothness of expert trajectories plays an important role. For instance, when substituting the initial non-momentum Stochastic Gradient Descent (SGD) with a superior optimizer, it does enhance the accuracy of expert model. However, this replacement markedly intensifies the rate of parameter changes in expert model, rendering it less smooth and moderate, ultimately resulting in difficulty of student parameter alignment and a significant decline in the distillation outcomes. 

Besides, the student parameter alignment exhibits a heightened susceptibility to two randomly initialized variables, namely the initial samples and the initial starting epochs of expert trajectories. The above two factors together also result in optimization instability of the distillation process. Particularly when distilling datasets of larger scale and higher resolution, it may lead to the generation of images akin to random noise and ``black holes''.

Finally, we argue that cumulative errors have a significant impact. It's not just the discrepancy between distillation and evaluation proposed by \cite{du2022minimizing} that results in cumulative errors. The long interval between inner and outer loop under the bi-level optimization structure will also generate accumulated errors, which degrade the final performance.

Building on the identified limitations, we propose Alignment with Smooth expert Trajectories (AST), which devises enhancements for both the expert trajectory generation and student parameter alignment phases, illustrated in \cref{fig:imtt}. Initially, we propose a combination of clipping loss and gradient penalty under a more proficient optimizer to modulate the parameter update rate, ensuring sustained high expert performance during the training of expert models. This refined trajectory, being smooth and high-quality, lays a solid foundation for optimizing student parameter alignment with expert. Secondly, we propose a method encompassing representative sample initialization and inner loop loss balancing, aimed at reducing sensitivity to random variable initialization and stabilizing the matching process. Finally, we propose two alleviation strategies: intermediate matching loss and expert model weight perturbation, devised to mitigate cumulative error propensity. Our proposed methods have been extensively evaluated on datasets varying in scale, size, and resolution, with results affirming their effectiveness, markedly outperforming state-of-the-art works.

\textbf{Contributions.} In summary, our main contributions can be summarized as follows:
\begin{itemize}
    \item We argue that the smoothness of expert trajectories significantly affects student parameter alignment. We propose incorporating clipping loss and gradient penalty to limit the rate of expert parameter change.
    \item For better alignment with expert trajectory, we propose representative sample initialization and inner loop loss balancing strategies to alleviate the impact of stochasticity from initialization variables.
    \item We propose two enhancement strategies, namely intermediate matching loss and the perturbation of expert model weights, to mitigate cumulative errors from various sources.
\end{itemize}

\section{Related Works}

\subsection{Dataset Distillation}
\label{sec:related}
The existing DD frameworks can be divided into four parts~\cite{sachdeva2023data}, namely Meta-model Learning, Gradient Matching, Distribution Matching, and Trajectory Matching.

Wang \textit{et al.} \cite{wang2018dataset} first introduces the dataset distillation problem and uses bi-level optimization similar to Model-Agnostic Meta-Learning (MAML) \cite{MAML}. Following the Meta-Learning frameworks, recent works include adding soft labels~\cite{bohdal2020flexible,sucholutsky2021soft}, developing closed-form approximation KIP~\cite{kip1,kip2} and FRePo~\cite{FRePo}, and applying prior towards the synthetic dataest~\cite{Factorization,cazenavette2023generalizing}. Gradient Matching conducts a single-step distance matching between the network trained on the original dataset and the identical network trained on synthetic data, methods including DC~\cite{GM}, DSA~\cite{dsa}, DCC~\cite{dcc} and IDC~\cite{idc}. To alleviate complex bi-level optimization and second-order derivative computation, Distribution Matching directly matches the distribution of original data and synthetic data with a single-level optimization, methods include DM~\cite{DM}, CAFE~\cite{cafe}, IT-GAN~\cite{it-gan}. However, the above-mentioned works are all short-range matching methods and may easily cause an overfitting problem~\cite{yu2023dataset}. 

Trajectory Matching \cite{mtt}, as a long-range framework, leverages the well-trained expert training trajectory as reliable experts and aims to minimize the distance between expert and student trajectories in the parameter alignment process of the second stage, which improves the performance of distilled dataset. Recent works include TESLA~\cite{tesla} that reparameterizes the computation of matching loss and FTD~\cite{du2022minimizing} that directly incorporates the calculated error term into the loss function used for buffer generation. Nevertheless, these works still have their own issues, such as ignoring the mutual influence between smoothness of expert trajectory and effectiveness of student parameter alignment, susceptibility to random initial variables, and cumulative errors from multiple sources.

\subsection{Relationship between Expert and Student}

Currently, there exist some methods in the field of Knowledge Distillation (KD) that explore the factors of what makes experts conducive to students learning or how to transfer more valuable knowledge from highly proficient experts. Huang \textit{et al.}~\cite{huang2022knowledge} has empirically pointed out that simply enhancing an expert's capabilities or employing a stronger expert can, paradoxically, result in a decrease in student performance. In some cases, this decline can even be more pronounced than if students were to begin training from scratch with vanilla KD. Consequently, they have proposed a loose matching approach to replace the traditional Kullback-Leibler (KL) divergence.

Similarly, Yuan \textit{et al.}~\cite{yuan2023student} argues that simplifying the expert model's outputs can offer greater advantages to the student's learning process. Meanwhile, Shao \textit{et al.}~\cite{shaoteaching} proposes the principle that ``experts should instruct on what ought to be taught' and introduces a data-based knowledge distillation method. All of the methods mentioned above underscore the intimate relationship between experts and students, necessitating a thorough exploration of how to enhance the abilities of both experts and students consistently. Nonetheless, there remains a vacancy in this relationship under the context of Dataset Distillation.


\section{Preliminaries}
\subsection{Problem Statement}




We introduce the problem statement of Dataset Distillation. Given a real dataset $\mathcal{D}_{real} = \{(x_i, y_i)\}_{i=1}^{|\mathcal{D}_{real}|}$, where the examples $x_i \in \mathbb{R}^d$ and the class labels $y_i \in \mathcal{Y} = \{1, 2, \ldots, C\}$ and $C$ is the number of classes. The essence of Dataset Distillation is to generate a distilled dataset $\mathcal{D}_{syn}$ with significantly fewer synthetic example-label pairs than $\mathcal{D}_{real}$. The goal is for a model $f$, when trained on $\mathcal{D}_{syn}$, to maintain the comparable performance of a counterpart trained on the original dataset $\mathcal{D}_{real}$.

The distilled dataset $\mathcal{D}_{syn}$ is defined as a set of pairs $\{(s_i, y_i)\}_{i=1}^{|\mathcal{D}_{syn}|}$, with $s_i$ as elements in $\mathbb{R}^d$ and $y_i$ belonging to $\mathcal{Y}$. For $\mathcal{D}_{syn}$, each class is represented by $ipc$ examples, thus giving $\mathcal{D}_{syn}$ a cardinality of $ipc \times C$, which is significantly lesser than that of $\mathcal{D}_{real}$ ($|\mathcal{D}_{syn}| \ll |\mathcal{D}_{real}|$). The optimal weight parameters, denoted as $\theta$, are those which minimize the empirical loss across the synthetic dataset $\mathcal{D}_{syn}$, formalized by the following equation:

\begin{equation}
\theta = \arg\min_{\theta} \sum_{(s_i,y_i) \in \mathcal{D}_{syn}} \ell(f_{\theta}, s_i, y_i)
\end{equation}
where $\ell$ can be an arbitrary loss function which is taken to be the cross entropy loss in this paper. The objective of Dataset Distillation is to construct an informative synthetic dataset $\mathcal{D}_{syn}$ that serves as an approximate solution to the optimization problem outlined below:

\begin{equation}
\begin{aligned}
\label{eq:base_arg1}
\mathcal{D}_{syn} &= \arg \min \mathcal{L}_{\mathcal{T}_{Test}}(f_{\theta}) \\
where \quad \mathcal{D}_{syn} &\subseteq \mathbb{R}^d \times \mathcal{Y}, |\mathcal{D}_{syn}| = ipc \times C
\end{aligned}
\end{equation}
To address the optimization challenge presented in \cref{eq:base_arg1}, Wang et al. \cite{wang2018dataset} employed a straightforward approach by substituting $\mathcal{T}_{Test}$ with $\mathcal{D}_{real}$ due to the unavailability of $\mathcal{T}_{Test}$ during the distillation phase. This forms the meta-learning based framework.

\subsection{Various Objective Functions of DD}
\textbf{Gradient-based Matching.} In addition to using the results on $\mathcal{D}_{real}$ as the objective function mentioned above, the gradient on real data can also be chosen as a reference. Under this, the objective function can be expressed as follows:

\begin{equation}
\arg \min_{\mathcal{D}_{syn}} \mathbb{E}_{\theta} \left[ \sum_{t=0}^{T} D\left( \nabla_{\theta} \mathcal{L}_{\mathcal{D}_{real}}(\theta_t), \nabla_{\theta} \mathcal{L}_{D_{\text{syn}}}(\theta_t) \right) \right]
\end{equation}
where $D(\cdot)$ is a distance metric such as cosine distance.

\textbf{Distribution-based Matching.} Distribution matching uses feature maps obtained on real data and distilled dataset as a proxy task, which can be expressed as:

\begin{equation}
\arg \min_{\mathcal{D}_{syn}} \mathbb{E}_{\psi} \left[ \left\| \mathbb{E}_{x_i \sim \mathcal{D}_{real}} [\psi(x_i)] - \mathbb{E}_{s_i \sim \mathcal{D}_{syn}} [\psi(s_i)] \right\|^2 \right]
\end{equation}
where $\psi(\cdot)$ refers to the process of extracting feature maps.

\textbf{Trajectory-based Matching.} The objective of matching expert trajectory is to align the training trajectory that is on $\mathcal{D}_{real}$ with those on $\mathcal{D}_{syn}$. This can be formulated as below:

\begin{equation}
\label{eq:tm_base}
\arg\! \min_{\mathcal{D}_{syn}} \mathbb{E}_{\theta} \! \left[ \sum_{t=0}^{T-\mathcal{M}} \!D\! \left( \theta_{t+\mathcal{M}}^{\mathcal{D}_{real}}, \theta_{t+\mathcal{N}}^{\mathcal{D}_{syn}} \right)\! \bigg/ \! D\! \left( \theta_{t+\mathcal{M}}^{\mathcal{D}_{real}}, \theta_{t}^{\mathcal{D}_{real}} \right) \right]
\end{equation}
similarly, $D(\cdot)$ is a distance metric to calculate the divergence of model parameters. $\mathcal{M}$ and $\mathcal{N}$ denote the number of intervals in the trajectories of the expert and the student, respectively. Further details are provided in \cref{interloss}.

\section{Design of AST}
\subsection{Overview}
As described earlier, we choose the long-range Matching Training Trajectories (MTT) as our framework. MTT uses expert training trajectories as a golden reference. Under this case, the expert trajectory is a means of high-level extracted information over the original large dataset. Then, the parameter of student model is aligned towards the expert trajectory and update the distilled dataset through backpropagation. However, through observation and analysis, we have still identified significant improvement space. By implementing a series of meticulous optimizations and providing more comprehensive guidance, the effectiveness of the distilled dataset and stability of distillation process can be substantially enhanced. Specifically, our proposed AST method encompasses three dimensions. In \cref{Generate Smooth Expert Trajectories}, we illustrate the mutual connection between expert and student and propose a way to maintain the expert trajectory smoothness under a more potent optimizer. Building on this improved trajectory, an enhanced parameter alignment method is proposed to balance the stochasticity (in \cref{Balance Stochasticity during Parameter Matching}) and alleviate the accumulated error (in \cref{Alleviate the Propensity for Accumulated Error}). The framework of AST is illustrated in \cref{fig:imtt} and our whole training algorithm can be found in \cref{sec:Training Algorithm}.

\subsection{Generate Smooth Expert Trajectories}
\label{Generate Smooth Expert Trajectories}


\subsubsection{Catastrophic Discrepancy with Stronger Expert Trajectory}
\label{catastrophic}

As depicted in \cref{intro}, the influence of an expert on DD has not been adequately investigated, especially as the performance of the expert model becomes stronger, for instance, through the use of better optimizers. With this regard, as \cref{tab:optimizer}, we demonstrate both results of the expert model and the distilled dataset achieved with various optimizers.

\begin{table}[h]
    \begin{center}
        \resizebox{0.75\linewidth}{!}{%
        \begin{tabular}{@{}ccc@{}}
        \toprule
        \textbf{Optimizer (Expert)} & \textbf{Acc. (Expert)} & \textbf{Acc. (Distill)} \\ \midrule
        Naïve SGD          & 48.6                & 39.7                    \\
        SGD w/ Mom         & 57.1                & 18.8                    \\
        ADAM               & 52.7                & 18.1                    \\ \bottomrule
        \end{tabular}%
        }
    \end{center}
    \captionsetup{skip=-3pt} 
    \caption{Expert model accuracy and distilled results on various expert trajectories using various optimizers on CIFAR-100 under IPC=10.}
    \label{tab:optimizer}
\end{table}

We have observed that when employing SGD with momentum or ADAM as optimizers, although they generate stronger expert trajectories, theoretically elevating the upper limit of distillation dataset performance, this increment does not directly connect with improved results of the distilled dataset. On the contrary, the accuracy of the distilled dataset on the validation set experiences a substantial decline, even falling behind the vanilla MTT method. This indicates that the knowledge encapsulated within the more potent expert trajectories is not effectively assimilated during following distillation. The underlying reason is that the utilization of more potent optimizers often incorporates the previously accumulated gradient into the optimization process, which facilitates faster convergence of expert model~\cite{goh2017why}. However, this causes the expert model's parameters to update too rapidly, surpassing the limitation of what can be distilled during the second stage. It is detrimental to the student parameter alignment. For specific experimental details, please refer to \cref{buffer_ana}. For theoretical analysis, please refer to \cref{sec:whypoor}.

Consequently, our proposed approach aims to enhance the performance of the expert model while simultaneously moderating the rate of parameter updates in the training process of expert models. Our strategy allows us to obtain a series of smooth and high-quality expert trajectories.

\begin{figure*}[t]
    \centering
    \includegraphics[width=1\linewidth]{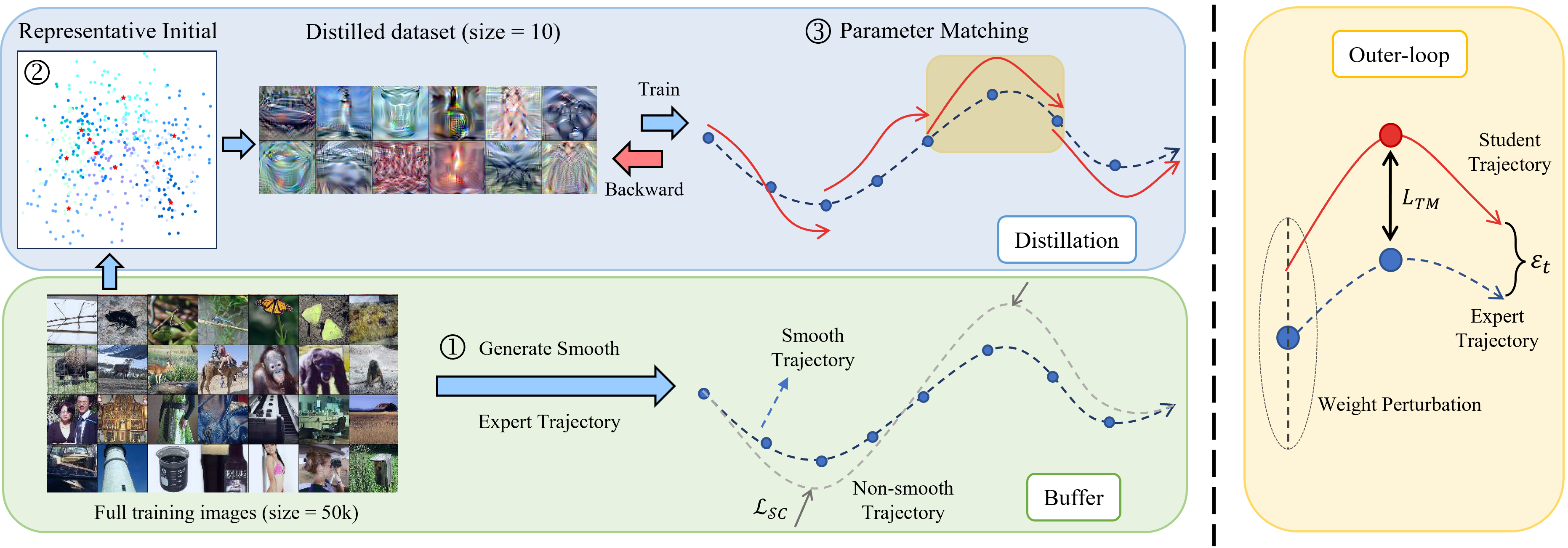}
    \caption{Illustration of AST method: (a) Buffer: We generate the smooth and high-quality expert trajectory (blue dash curve) on the real dataset $\mathcal{D}_{real}$. The grey dash curve is a much steeper expert trajectory without applying $\mathcal{L}_{SC}$. (b) Distillation: We first select representative initialization samples from the original training images. Then, the synthetic dataset $\mathcal{D}_{syn}$ is optimized to match the segments of the expert trajectory through parameter alignment. We apply a balanced inner-loop loss $\hat{\ell}_{BIL}$ to mitigate the influence of random initialized expert starting epoch $t$. The red solid curve denotes student trajectory. (c) Outer-loop in distillation: We show how student trajectories are matched with expert trajectories within a single iteration. We introduce weight perturbation $\mathbf{d}_{l, j}$ to the well-trained expert model parameters. Subsequently, we calculate the intermediate matching loss $\mathcal{L}_{TM}$ in the middle of the loop. These two methods can help mitigate the accumulated errors $\varepsilon_t$.}
    \label{fig:imtt}
\end{figure*}

\subsubsection{Why do Stronger Expert Trajectories Perform Even Poorly?}
\label{sec:whypoor}
We hope to theoretically explain why stronger expert trajectories cannot contribute to consistent improvement of distilled datasets or even cause model collapse. Here, the stronger expert trajectories refer to using ADAM or SGD with momentum as an optimizer. Applying potent optimizers will accelerate the convergence due to the momentum. ADAM uses both first and second-order momentum. We take SGD with momentum as an example. 

In the buffer phase, our objective function is as follows:
\begin{equation}
\begin{aligned}
\label{eq:bufferloss}
    \theta^*_t &= \underset{\theta_t}{\arg \min } \quad \mathcal{L}(\theta_t^*,x,y) \\
    \text{where } \mathcal{L}(\theta_t^*,x,y) &= \frac{1}{\left| \mathcal{D}_{real}\right|  }\sum_{(x,y)  \in \mathcal{D}_{real}}\!\left[ y\log(\theta^*_{t}(x))\right]
\end{aligned}
\end{equation}
in which $\theta^*_t$ denotes the expert model at the $t$ epoch. Then, we depict the optimization process:
\begin{equation}
\begin{aligned}
        g_t & =\nabla_\theta \mathcal{L}(\theta^*_t)=\frac{\partial \mathcal{L}}{\partial(\theta^*_t)} \\
        \theta_{t+1}^* & =\theta_t^*-v_t
\end{aligned}
\end{equation}
in which $v_t$ represents the momentum:
\begin{equation}
\begin{aligned}
\label{eq:vt}
v_t &= \gamma \cdot v_{t-1} + \eta \cdot g_t\\
\end{aligned}
\end{equation}
here, $\eta$ is the learning rate and $\gamma$ is the momentum factor. $v_t$ is the velocity update that contains the current gradient $g_t$ and the exponential moving average of all the past time's gradients. We expand the \cref{eq:vt} explicitly:
\begin{equation}
\begin{aligned}
v_t & =\eta \cdot g_t+\gamma \cdot v_{t-1} \\
& =\eta \cdot g_t+\gamma\left[\eta \cdot g_{t-1}+\gamma \cdot v_{t-2}\right] \\
& \cdots \\
& =\eta \cdot\left[\sum_{k=1}^{t-1} \gamma^{t-k} \cdot g_k+g_t\right]
\end{aligned}
\end{equation}

Compared to the normal SGD without momentum used in all the prior works, each update contains a cumulative item:
\begin{equation}
\label{eq:delta}
    \delta_t = v_t - \eta \cdot g_t = \sum_{k=1}^{t-1} \gamma^{t-k} \cdot g_k
\end{equation}
This cumulative term $\delta_t$ can help the model converge faster during the training of expert models, avoiding repeated oscillations at saddle points. However, during the second stage of distillation, it may cause significant errors. The gradient's moving average term $\delta_t$ contains a lot of redundant information, which will increase the difficulty of student parameters alignment.

In the distillation process, we only take a segment of the expert trajectories. Assume the expert starting point is $t$ and the endpoint is $t+\xi_i$. We use the expert model $\theta^*_t$ to initialize the student network $\hat{\theta_t}$. We can then formulate the expert parameter changing as follows:
\begin{equation}
\begin{aligned}
    \theta^*_{t+1} &= \theta^*_{t} - v_{t-1}\\ 
    \theta^*_{t+2} &= \theta^*_{t+1} - v_{t}\\ 
    & \cdots \\
    \theta^*_{t+\xi_i} &= \theta^*_{t+\xi_{i}-1} - v_{t+\xi_{i}-2} \\
\end{aligned}
\end{equation}
After sum up, we can get
\begin{equation}
\begin{aligned}
\label{eq:12}
    \theta^*_{t+\xi_i} - \theta^*_{t} &= \sum_{p=t-1}^{t+\xi_{i}-2} v_{p}
\end{aligned}
\end{equation}
We take the \cref{eq:delta} into the former equation so we get the extra item that needs to be alignment during the distillation:
\begin{equation}
\begin{aligned}
    \Delta^*_{t-1} = \sum_{p=t-1}^{t+\xi_{i}-2} \delta_p = \sum_{p=t-1}^{t+\xi_{i}-2} \quad \left[\sum_{k=1}^{p-1} \gamma^{p-k} \cdot g_k\right]
\end{aligned}
\end{equation}

However, the gradient accumulation term used in the expert model during distillation is a kind of truncated accumulation term in the second phase, whose starting point already includes the momentum of $v_{t-1}$. But for the student model, the gradient accumulation starts from the zero origin and ends at $\mathcal{N}$, as shown in \cref{alg:DD} line 8-15. The origin does not have the initial gradient accumulation term. This leads to inconsistencies in parameter alignment, even if we also incorporate momentum during the distillation process. At the same time, this inconsistency will continue to amplify with iterations, resulting in difficulties in final convergence. The following formula represents the parameter updates in the student network: 
\begin{equation}
\begin{aligned}
    \hat{\theta}_{t+1} &= \hat{\theta}_{t} - v'_{0}\\ 
    \hat{\theta}_{t+2} &= \hat{\theta}_{t+1} - v'_{1}\\ 
    & \cdots \\
    \hat{\theta}_{t+n} &= \hat{\theta}_{t+n-1} - v'_{n} \\
\end{aligned}
\end{equation}
After sum up, we can get
\begin{equation}
\begin{aligned}
\label{eq:15}
    \hat{\theta}_{t+n} - \hat{\theta}_{t} &= \sum_{p=0}^{n-1} v'_{p}
\end{aligned}
\end{equation}
Similarly, We take the \cref{eq:delta} into the former equation so we get the extra gradient moving average item for the student model during the distillation:
\begin{equation}
\begin{aligned}
    \hat{\Delta}_{0} = \sum_{p=0}^{n-1} \delta_p' = \sum_{p=0}^{n-1} \quad \left[\sum_{k=1}^{p-1} \gamma^{p-k} \cdot g_k\right]
\end{aligned}
\end{equation}

When we calculate the numerator of alignment loss in \cref{eq:tm_base}, it is equivalent to calculating the difference between \cref{eq:12} and \cref{eq:15}.
\begin{equation}
\label{eq:17}
    \left\|\hat{\theta}_{t+n}-\theta_{t+\xi_i}^{*}\right\|_{2}^{2} = \left\| \sum_{p=0}^{n-1} v'_{p} - \sum_{p=t-1}^{t+\xi_{i}-2} v_{p}\right\|_{2}^{2}
\end{equation}
and the additional gradient average term inside \cref{eq:17} that needs to be aligned is:
\begin{equation}
\begin{aligned}
    \varepsilon_t &= \left\| \Delta^*_{t-1} - \hat{\Delta}_{0} \right\|_{2}^{2}
\end{aligned}
\end{equation}

The above has demonstrated the errors $\varepsilon_t$ introduced by incorporating momentum into the optimizer. The source of these errors is twofold. On one hand, the student trajectory, to be aligned in the distillation phase, is only one truncated segment from $\hat{\theta}_t$ to $\hat{\theta}_{t+n}$ initialized by a starting point of expert trajectory $\theta^*_t$. The student network lacks the initial gradient accumulation term, resulting in a divergence gap at the beginning. On the other hand, the presence of momentum during iterations exacerbates and magnifies the inherent inconsistencies, leading to further amplified errors. These two factors contribute to the challenge of student parameters alignment, ultimately failing to consistently improve distillation outcomes from better expert trajectories. The clipping loss and gradient penalty we propose in the next section can restrict the gradient accumulation term, keeping it within a smaller range from beginning to end. Specifically, we generate a series of smooth and high-quality expert trajectories, where the changing of parameters exhibits a flat trend while expert performance maintains improvement.


\subsubsection{Better Expert Trajectory under Smoothness Constraint}
\label{gensmoothexp}

To constrain the changing speed of expert parameters, the straightforward method involves the utilization of value clipping on the cross-entropy loss, with a specific focus on the initial epochs. Gradually, the clipping coefficient is incremented until it reaches a value of 1, which is then held constant. Our findings indicate that although loss clipping may impact the final expert performance to some extent, its key benefit lies in mitigating the rapid changes in model parameters, particularly during the initial few epochs. 

Besides, to maintain the overall smoothness of the expert trajectory, we introduce the concept of gradient penalty, building upon the insights from the Wasserstein GAN (WGAN)~\cite{wgan,wgan-gp}. The goal of employing gradient penalty is to force the model to satisfy Lipschitz continuity, preventing abrupt and erratic changes in the model's parameters. We incorporate gradient penalty by adding a regularization term to the loss function. This term is formulated as the squared norm of the gradient $\nabla_{x_i} \mathcal{W}(x_i)$ of the model's output with respect to its input. By penalizing large gradients, we incentivize the model to produce outputs that change gradually as inputs vary slightly. This has the effect of maintaining a smoother transition between data points and preventing sudden shifts that could lead to instability. We formulate the final loss equation as below:

\vspace{-0.5cm}
\begin{equation}
\label{eq:l_sc}
\mathcal{L}_{SC}\!=\!\underbrace{\lambda \log \frac{\exp \left(x_{i, y_{i}}\right)}{\sum_{c=1}^{C} \exp \left(x_{i, c}\right)}}_{\text {Clipped CELoss}}+\underbrace{\mu\!\underset{x_i \sim \mathbb{P}_{x}}{\mathbb{E}}\left[\left(\left\|\nabla_{x_i} \mathcal{W}(x_i)\right\|_{2}\!-\!\mathcal{K}\right)^{2}\right]}_{\text {Gradient Penalty}} 
\end{equation}
where $\left\|\nabla_{x_i} \mathcal{W}(x_i)\right\|_2$ represents a dual-sided penalty that aims to constrain the gradient norm to values below a predetermined threshold denoted as $\mathcal{K}$. Typically, $\mathcal{K}$ is set at 1. The coefficient $\lambda$ operates within a range of 0.5 to 1 during the initial few epochs, while the coefficient $\mu$ is responsible for scaling the gradient penalty effect.

\subsection{Balancing Stochasticity from Initial Variables}
\label{Balance Stochasticity during Parameter Matching}

\subsubsection{Representative Initial for Synthetic Dataset}
\label{repinitial}

Many existing dataset distillation algorithms initialize $\mathcal{D}_{syn}$ either by employing random initialization with Gaussian noise or by randomly selecting a subset of images from the real dataset. However, this approach can inadvertently introduce outliers, lead to the inclusion of samples with similar features, or overlook certain aspects of the feature space, which may introduce huge biases and stochasticity. Moreover, as distilled datasets often end up with a limited number of samples per class, it becomes crucial to efficiently encapsulate the inherent information of the original dataset within these limited samples. To mitigate these limitations, we propose representative initial samples for $\mathcal{D}_{syn}$.

Leveraging the benefits of the MTT framework can simplify the process of selecting representative initialization samples for $\mathcal{D}_{syn}$.
In the buffer stage, we have already obtained the training trajectory of the expert model. Then, we can acquire a well-trained model easily by loading the parameters of expert trajectory. Next, we input all real data of the same class into the model, obtaining feature vectors before entering the fully connected layer. Subsequently, we perform clustering on these feature maps, utilizing the K-Means algorithm to partition them into multiple sub-clusters.
The value of K is chosen based on the desired number of distilled samples. These sub-cluster centroids are then selected as exemplary initialization samples. 
The primary objective is to enhance the initialization process by strategically selecting representative samples that are better suited for distillation.

\subsubsection{Balanced Inner-loop Loss}
\label{balanceloss}

\begin{figure}[t]
    \centering
    \begin{subfigure}{0.48\linewidth}
        \centering
        \includegraphics[width=\linewidth]{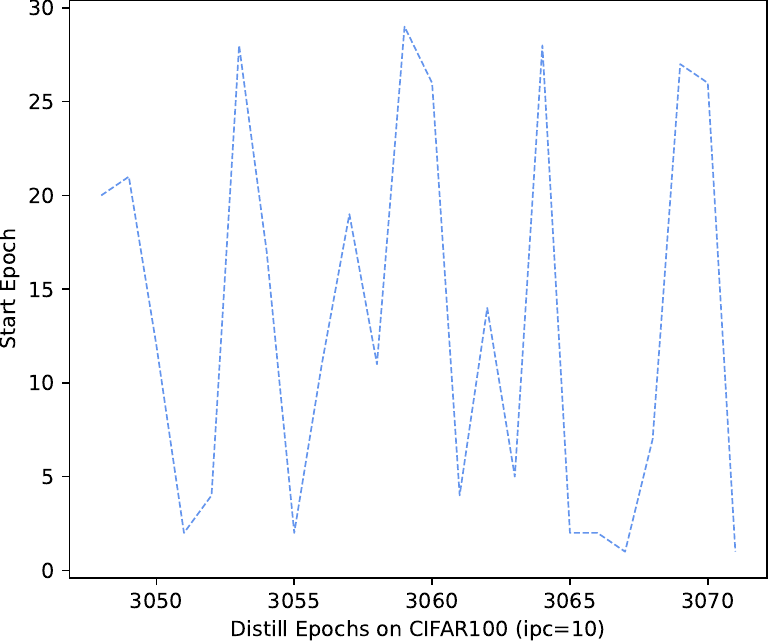}
        \caption{Randomly Start Epoch}
        \label{fig:startepoch}
    \end{subfigure}
    \begin{subfigure}{0.48\linewidth}
        \centering
        \includegraphics[width=\linewidth]{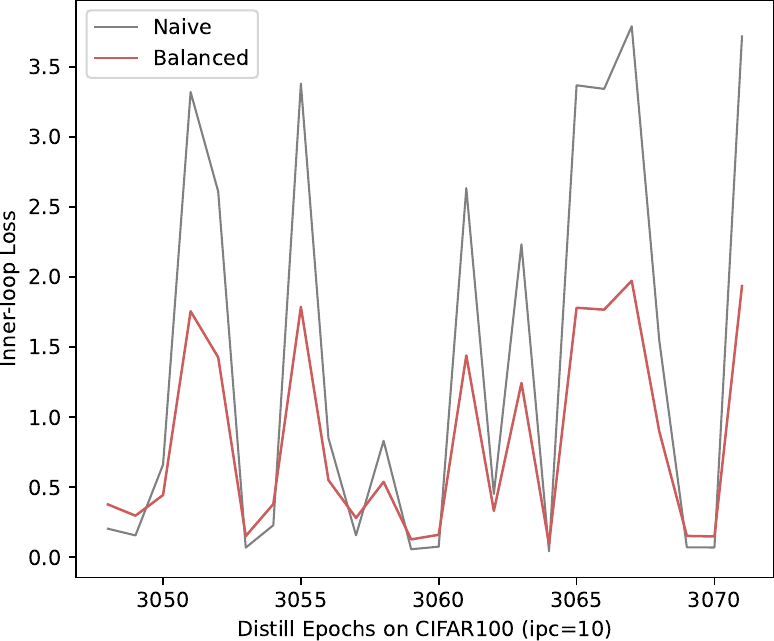}
        \caption{Inner-loop Loss}
        \label{fig:balance}
    \end{subfigure}
    \caption{Because of the random variable of starting epoch for each iteration (as depicted in the blue dashed plot in (a)), it leads to substantial fluctuations in the loss scale within the inner loop, as indicated by the solid grey plot in (b). After employing a balanced inner-loop loss, the extent of loss variation is constrained to a lesser degree during the early starting epochs, while allocating more weight to the later starting epochs, as illustrated by the red plot in (b).}
    \label{fig:BIL}
\end{figure}

During the second stage, student parameter alignment exhibits sensitivity to the randomly initialized expert starting epochs $t$ in each iteration. This sensitivity is specifically manifested in significant fluctuations of the inner-loop loss $\ell_{IL}$. Depending on the various starting points shown in \cref{fig:startepoch}, the largest loss is nearly 60 times greater than the smallest loss shown in \cref{fig:balance}. The fluctuations in the inner-loop loss introduce a notable level of instability into the distillation procedure, which impedes the parameters' alignment and hinders the consistent acquisition of accurate distilled data.

\begin{table}[t]
    \begin{center}
        \resizebox{0.75\linewidth}{!}{%
        \begin{tabular}{@{}ccc@{}}
        \toprule
        \textbf{Initial Samples} & \textbf{Starting Epoch} & \textbf{Acc. (Distill)} \\ \midrule
        Random                   & Random                  & 39.7                    \\
        Representative           & Random                  & 40.3                    \\
        Random                   & Balanced Strategy       & 40.1                    \\ \bottomrule
        \end{tabular}%
        }
    \end{center}
    \captionsetup{skip=-3pt} 
    \caption{After using the stochasticity balancing strategy during parameter matching, there has been a consistent improvement in the performance of the distilled dataset.}
    \label{tab:randomvar}
\end{table}

One naive method is to discard the practice of entirely random selection for the starting epoch and instead opt for a method that references the preceding starting point and selects the subsequent starting point within a reasonable range. However, experiments (details in \cref{Impact of Balance Strategy}) have demonstrated that this dynamic selection approach fails to enhance the final performance. We speculate that employing a range selection may introduce potential bias into the distillation process, which in turn causes the model to learn incorrect inductive bias. Hence, we propose to balance the loss within the inner loop. 

When the starting epoch $t$ is much smaller, which means the network is under convergence, it will typically generate a much larger loss $\ell_{IL}$ leading the gradient descent towards this direction. Thus, we add a penalty coefficient $\nu$ to narrow the loss, clipping the loss to a small extent. 
Conversely, when the starting epoch $t$ is large (close to $\mathcal{T_{+}}$), the optimized steps become much smaller due to the decreased loss. We also add a coefficient $\nu$ to encourage a much bigger gradient.
Under this circumstance, the balanced inner-loop loss $\hat{\ell}_{BIL}$ can be formulated as:
\begin{equation}
\label{eq:nu}
\begin{aligned}
    \nu=\left\{\begin{matrix}
     log(|start-middle|+\vartheta ),& start\ge middle \\ 
     1/log(|middle-start|+\vartheta ),& start<middle
    \end{matrix}\right. \\ \nonumber
\end{aligned}
\end{equation}
\vspace{-0.88cm}
\begin{equation}
    \hat{\ell}_{BIL} = \nu \times \ell_{IL}
\end{equation}
where $start$ represents the starting epoch, $middle$ refers to half of the maximum starting epoch and $\vartheta$ is included to balance the log scale and prevent negative values. As shown in \cref{fig:BIL}, using balanced loss $\hat{\ell}_{BIL}$ can reduce the loss fluctuation in the inner loop. For $\ell_{IL}$, it is similar to cross-entropy loss:

\begin{equation}
\label{eq:celoss}
    \ell_{IL} =  \frac{1}{\left| \mathcal{D}_{syn}\right|  }\sum_{(s_i,y_i)  \in \mathcal{D}_{syn}}\!\left[ y_i\log(\hat{\theta}_{t}(\mathcal{A}(s_i)))\right]
\end{equation}
where $\mathcal{A}$ represents the differentiable siamese augmentation~\cite{dsa}, while $\hat{\theta}_{t}$ denotes the student network parameterized using the expert's $t$-th epoch.network, $t$ is smaller than the defined maximum start epoch $\mathcal{T_{+}}$ and $\mathcal{D}_{syn}$ is the distilled samples. From \cref{tab:randomvar}, a consistent improvement of the distilled dataset is observed after using the stochasticity balancing strategy.

\begin{algorithm}[h]
\caption{Dataset Distillation in Expert Trajectory Generation}
\label{alg:buffer}
\KwIn{$\mathcal{D}_{real}$: real dataset.}
\KwIn{$f$: expert network with weights $\theta$.} 
\KwIn{$\eta$: learning rate.}
\KwIn{$E$: training epochs.}
\KwIn{$T$: iterations per epoch.}

\For {$e = 1$ to $E$}{
    \For{$t = 1$ to $T$, mini-batch $\mathcal{B} \subset \mathcal{D}_{real}$}{
        Compute loss $\mathcal{L}$ based on \cref{eq:l_sc} \;
        Compute gradients: $g_\mathcal{L}=\nabla_\theta \mathcal{L}_{\mathcal{B}}\left(f_{\theta_t}\right)$ \;
        Compute the momentum: $v_t \!= \gamma \cdot v_{t-1}\! + \eta \cdot g_{\mathcal{L}}$ \;
        Update weights: $\theta_{t+1} \leftarrow \theta_t - v_t$ \;
    }
    Record weights $\theta_T$ of expert trajectory at the end of each epoch 
}

\KwOut{Smooth expert trajectory $\tau :=  \{\theta_T\}$}
\end{algorithm}

\subsection{Alleviate the Propensity for Accumulated Error}
\label{Alleviate the Propensity for Accumulated Error}

\subsubsection{Intermediate Matching Loss}
\label{interloss}

Trajectory matching is a long-range framework, typically involving a larger number of steps denoted as $\mathcal{N}$ to match a smaller number of epochs denoted as $\mathcal{M}$. Here, $\mathcal{M}$ signifies the number of intervals between the starting and target point of the expert trajectory, while $\mathcal{N}$ represents the training steps of the student model on the distilled dataset. $\mathcal{N}$ is also equivalent to the number of iterations in the inner loop. After every $\mathcal{N}$ steps, it will execute a matching interaction with the expert trajectory.
However, extensive intervals of interaction can lead to the inner loop deviating from the correct direction prematurely. It will result in the accumulation of errors without sufficient and timely supervision guidance.
Therefore, we propose intermediate matching loss shown in \cref{fig:imtt} ``Outer-loop'', which optionally performs an additional parameter matching step within the inner loop. 

Specifically, we establish a set of intermediate matching points denoted as $\xi$, which consists of two parameters, $\mathcal{M}$ and $\mathcal{N}$. The matching points are selected on an average basis and can be represented as $\{\xi \} = \{ \left\lfloor  \mathcal{N} / \mathcal{M} \ \right\rfloor, \left\lfloor 2\!\times\!\mathcal{N} / \mathcal{M}\ \right\rfloor, ..., \left\lfloor (\mathcal{M}\!-\!1)\!\times\!\mathcal{N} / \mathcal{M}\ \right\rfloor\}$, where $\left\lfloor \cdot \right\rfloor$ is floor function. When the inner loop $n$ ($from~ 1~ to~ \mathcal{N}$) precisely aligns with ${\xi_{i} \in \{\xi\} }$, the intermediate matching loss function is computed as:
\vspace{-0.1cm}
\begin{equation}
\label{eq:L_TM}
    \mathcal{L}_{TM}=\frac{\left\|\hat{\theta}_{t+n}-\theta_{t+\xi_i}^{*}\right\|_{2}^{2}}{\left\|\theta_{t}^{*}-\theta_{t+\xi_i}^{*}\right\|_{2}^{2}}  
\end{equation}
in which $\theta_{t}^{*}$ is the expert training trajectory at randomly starting epoch $t$, also known as the initial expert weights. Starting from $\theta_{t}^{*}$, $\theta_{t+\xi_i}^{*}$ denotes $\{\xi\}$ steps trained by the expert network on real data and $\hat{\theta}_{t+n}$ stands for $n$ steps ($n<\mathcal{N}$) by student network trained on synthetic data correspondingly. The goal is to minimize the divergence between $\hat{\theta}_{t+n}$ and $\theta_{t+\xi_i}^{*}$, and $\left\|\theta_{t}^{*}-\theta_{t+\xi_i}^{*}\right\|_{2}^{2}$ is used to self-calibrate the magnitude across different iterations.
$\mathcal{L}_{TM}$ is then utilized to update the student parameters. 

\begin{algorithm}
\caption{Dataset Distillation in Student Parameter Alignment}
\label{alg:DD}
\KwIn{$\{\tau_i\}$: set of smoothed expert parameter trajectories trained on real dataset $\mathcal{D}_{real}$.} 
\KwIn{$\mathcal{M}$: number of intervals between starting and target expert trajectory.}
\KwIn{$\mathcal{N}$: distillation steps of student network.} 
\KwIn{$\mathcal{A}$: differentiable siamese augmentation.} 
\KwIn{$\mathcal{T_{+}}$: maximum start epoch.}
\KwIn{$\{\xi \} = \{ \left\lfloor  \mathcal{N} / \mathcal{M} \ \right\rfloor, ..., \left\lfloor (\mathcal{M}\!-\!1)\!\times\!\mathcal{N} / \mathcal{M}\ \right\rfloor\}$: set of intermediate matching epochs.}

Select representative initialization samples $\mathcal{D}_{syn} \sim \mathcal{D}_{real}$;  \tcp{Method~\ref{repinitial}}
Initialize trainable learning rate $\alpha := \alpha_0$\;
\For {\textbf{each} distillation step}{
    Sample smooth expert trajectory $\tau \sim \{\tau_i\}$ with $\tau :=  \{\theta_t\}$\;
    Choose random start epoch, $t \leqslant \mathcal{T_{+}}$\;
    Perturb weight on initial expert model with $\tau^* :=  \{\theta_t^*\}$; \tcp{ Method~\ref{perturb}}
    Initialize student network with expert parameters  $\hat{\theta}_t := \theta_t^*$\;
    \For{n = 1 to $\mathcal{N}$}{
        Sample a mini-batch of distilled images: $b_{t+n} \sim \mathcal{D}_{syn}$\;
        Compute the cross-entropy loss based on \cref{eq:celoss}. \;
        Get $\nu$ based on \cref{eq:nu} and balance $\ell_{IL}$: 
        $\hat{\ell}_{BIL} = \nu \times \ell_{IL}$; \tcp{Method~\ref{balanceloss}} 
        
        
        Update student model: $\hat{\theta}_{t+n+1}\!=\!\hat{\theta}_{t+n} \!-\! \alpha \nabla\hat{\ell}_{BIL}$\;
    \If{n \textbf{in} \{$\xi$\}}{
    Calculate intermediate matching loss $\mathcal{L}_{i}\!=\!\left\|\hat{\theta}_{t+n}\!-\!\theta_{t+\xi_i}^{*}\right\|_{2}^{2} /\left\|\theta_{t}^{*}\!-\!\theta_{t+\xi_i}^{*}\right\|_{2}^{2}$; \tcp{Method~\ref{interloss}}
    }
  }
  Get the final loss $\hat{\mathcal{L}} = \sum_{\xi_i} \beta_{i} \times \mathcal{L}_{i}$ \;
  Update $\mathcal{D}_{syn}$ and $\alpha$ with respect to $\hat{\mathcal{L}}$\;
    
}

\KwOut{Distilled dataset $\mathcal{D}_{syn}$ and learning rate $\alpha$}
\end{algorithm}

\subsubsection{Weight Perturbation on Initial Expert Model}
\label{perturb}
Another potential source of cumulative error stems from the disparity between the distillation and evaluation, as pointed out in~\cite{du2022minimizing}. Due to the discrete nature of the distillation process versus the continuous nature of the evaluation, cumulative errors can thus arise. Therefore, unlike incorporating the calculated error term into the buffer loss function \cite{du2022minimizing}, we introduce a simpler approach to simulate the potential expert weight deviation during the parameter alignment, named weight perturbation. The perturbation value can be calculated as follows:
\vspace{-0.1cm}
\begin{equation}
\begin{aligned}
    \mathbf{d}_{l, j} &= \frac{\mathbf{d}_{l, j}}{\left\|\mathbf{d}_{l, j}\right\|_{F}}\left\|\mathbf{\theta}_{l, j}\right\|_{F} \\
    \theta^{*}_{t} &= \theta_{t} + \rho * \mathbf{d}_{l,j} 
\end{aligned}
\end{equation}
where $\mathbf{d}_{l, j}$ is sampled from a Gaussian distribution $N(0,\,1)$ with dimensions same as $\mathbf{\theta}_{l,j}$. $\mathbf{d}_{l, j}$ is the $j$-th filter at the $l$-th layer of $\mathbf{d}$ and $\left \| \cdot  \right \|_F$ refers to the Frobenius norm. 
Finally, a coefficient $\rho$ is added to obtain the final $\theta^{*}_{t}$.

\subsection{Training Algorithm}
\label{sec:Training Algorithm}
We show the algorithm of expert trajectory generation under smoothness constraint in \cref{alg:buffer}.
We incorporate the methods in \cref{Balance Stochasticity during Parameter Matching} and \cref{Alleviate the Propensity for Accumulated Error} into the student parameter alignment stage and depict our detailed algorithm in ~\cref{alg:DD}.

\section{Experiments}

\subsection{Experiments Setup}
\label{Experiments_Setup}
The majority of our experimental procedures closely follow the previous works~\cite{mtt,tesla,du2022minimizing}, which ensure a fair and equitable basis for comparison. Each of our experiments comprises three essential phases: buffer (generating expert trajectories), distillation (student parameters alignment), and evaluation phase.
First, we generate 50 distinct expert training trajectories, with each trajectory encompassing 50 training epochs. Second, we synthesize a small distilled dataset (e.g., 10 images per class) from a given large real training set. Finally, we employ this learned distilled dataset to train randomly initialized neural networks and assess the performance of these trained networks on the real test dataset. 

\textbf{Datasets.} We verify the effectiveness of our method on both low- and high- resolution datasets distillation benchmarks, including CIFAR-10 \& CIFAR-100~\cite{krizhevsky2009learning}, Tiny ImageNet~\cite{le2015tiny} and ImageNet subsets~\cite{mtt}. Top-1 accuracy is reported to show the performance.

\begin{table*}[ht]
\begin{center}
\scriptsize
\renewcommand\arraystretch{0.72}
\resizebox{\linewidth}{!}{
\begin{tabular}{@{}c|ccc|ccc|cc@{}}
\toprule
\multirow{2}{*}{IPC} & \multicolumn{3}{c|}{\textbf{CIFAR-10}}                        & \multicolumn{3}{c|}{\textbf{CIFAR-100}}                       & \multicolumn{2}{c}{\textbf{Tiny ImageNet}} \\
                     & 1        & 10                & \multicolumn{1}{c|}{50}       & 1        & 10                & \multicolumn{1}{c|}{50}       & 1                    & 10                  \\ \midrule
Full Dataset         & \multicolumn{3}{c|}{84.8±0.1}                                & \multicolumn{3}{c|}{56.2±0.3}                                & \multicolumn{2}{c}{39.5±0.4}               \\ \midrule
Random       & 14.4±2.0 & 26.0±1.2 & 43.4±1.0 & 4.2±0.3 & 14.6±0.5     & 30.0±0.4 & 1.4±0.1          & 5.0±0.2          \\
Herding~\cite{chen2012super}      & 21.5±1.2 & 31.6±0.7 & 40.4±0.6 & 8.4±0.3 & 17.3±0.3     & 33.7±0.5 & 2.8±0.2          & 6.3±0.2          \\
Forgetting~\cite{Forgetting}   & 13.5±1.2 & 23.3±1.0 & 23.3±1.1 & 4.5±0.2 & 15.1±0.3 & 30.5±0.3 & 1.6±0.1          & 5.1±0.2          \\ \midrule
DC~\cite{GM}                   & 28.3±0.5 & 44.9±0.5          & \multicolumn{1}{c|}{53.9±0.5} & 12.8±0.3 & 25.2±0.3          & \multicolumn{1}{c|}{-}        & -                    & -                   \\
DM~\cite{DM}                   & 26.0±0.8 & 48.9±0.6          & \multicolumn{1}{c|}{63.0±0.4} & 11.4±0.3 & 29.7±0.3          & \multicolumn{1}{c|}{43.6±0.4} & 3.9±0.2              & 12.9±0.4            \\
DSA~\cite{dsa}                  & 28.8±0.7 & 52.1±0.5          & \multicolumn{1}{c|}{60.6±0.5} & 13.9±0.3 & 32.3±0.3          & \multicolumn{1}{c|}{42.8±0.4} & -                    & -                   \\
CAFE~\cite{cafe}                 & 30.3±1.1 & 46.3±0.6          & \multicolumn{1}{c|}{55.5±0.6} & 12.9±0.3 & 27.8±0.3          & \multicolumn{1}{c|}{37.9±0.3} & -                    & -                   \\
FRePo~\cite{FRePo}      & 45.1±0.5 & 59.1±0.3 & 69.6±0.4 &25.9±0.1$^{\dag}$ &40.9±0.1 & - & 13.5±0.1$^{\dag}$ & 20.4±0.1 \\ \midrule
MTT~\cite{mtt}                  & 46.2±0.8 & 65.4±0.7          & \multicolumn{1}{c|}{71.6±0.2} & 24.3±0.3 & 39.7±0.4          & \multicolumn{1}{c|}{47.7±0.2} & 8.8±0.3              & 23.2±0.2            \\
TESLA~\cite{tesla}                & 48.5±0.8$^{\dag}$ & 66.4±0.8          & \multicolumn{1}{c|}{72.6±0.7} & 24.8±0.4 & 41.7±0.3          & \multicolumn{1}{c|}{47.9±0.3} & 9.8±0.4                    & 24.4±0.6                   \\
FTD~\cite{du2022minimizing}                  & 46.8±0.3 & 66.6±0.3$^{\dag}$          & \multicolumn{1}{c|}{73.8±0.2$^{\dag}$} & 25.2±0.2 & 43.4±0.3$^{\dag}$          & \multicolumn{1}{c|}{50.7±0.3$^{\dag}$} & 10.4±0.3             & 24.5±0.2$^{\dag}$            \\ \midrule
\textbf{Ours}                 & \textbf{48.8±0.9}         & \textbf{67.1±0.4} & \multicolumn{1}{c|}{\textbf{74.6±0.5}}         & \textbf{26.6±0.4}         & \textbf{44.4±0.6} &  \multicolumn{1}{c|}{\textbf{51.7±0.7}}         & \textbf{13.7±1.4}                     &  \textbf{25.7±1.1}                   \\ \bottomrule
\end{tabular}
}
\end{center}
\caption{Performance comparision trained with 128 width-ConvNet~\cite{convnet} to other state-of-the-art methods on the CIFAR and Tiny ImageNet. We cite the reported results from Sachdeva \textit{et al.}~\cite{sachdeva2023data} and Du \textit{et al.}~\cite{du2022minimizing}. IPC: Images Per Class. Bold digits represent the best results and $\dag$ refers to the second-best results among all the methods..}
\label{tab:main-result}
\end{table*}

\begin{figure*}[h]
    \centering
    \begin{subfigure}{0.32\linewidth}
        \centering
        \includegraphics[width=1\linewidth]{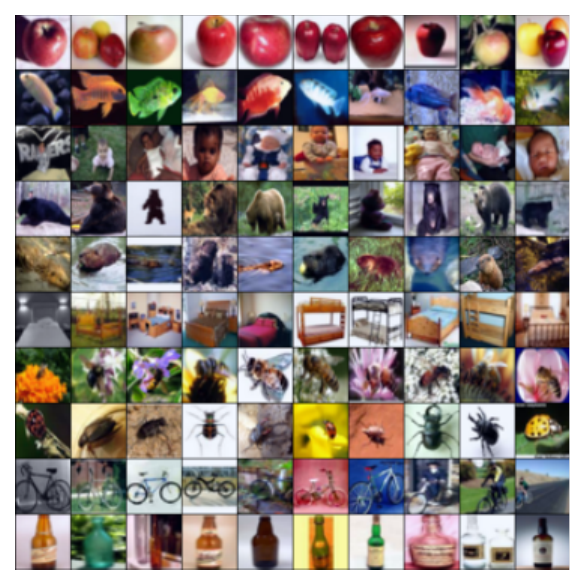}
        \caption{Original CIFAR100 images.}
        \label{origin}
    \end{subfigure}
    \begin{subfigure}{0.32\linewidth}
        \centering
        \includegraphics[width=1\linewidth]{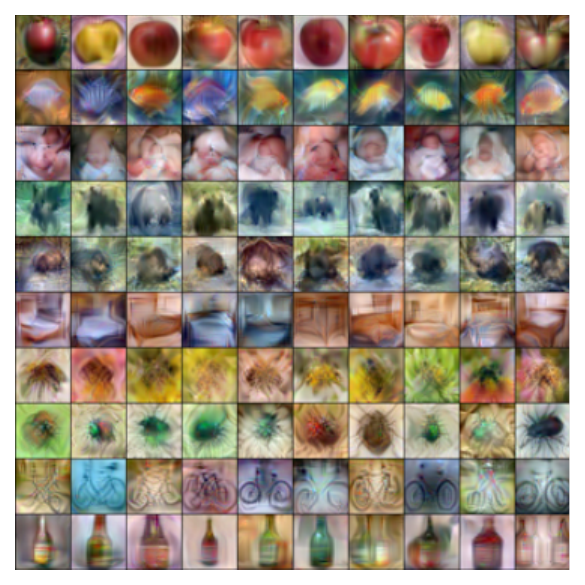}
        \caption{The synthetic images of MTT.}
        \label{MTT}
    \end{subfigure}
    \begin{subfigure}{0.32\linewidth}
        \centering
        \includegraphics[width=1\linewidth]{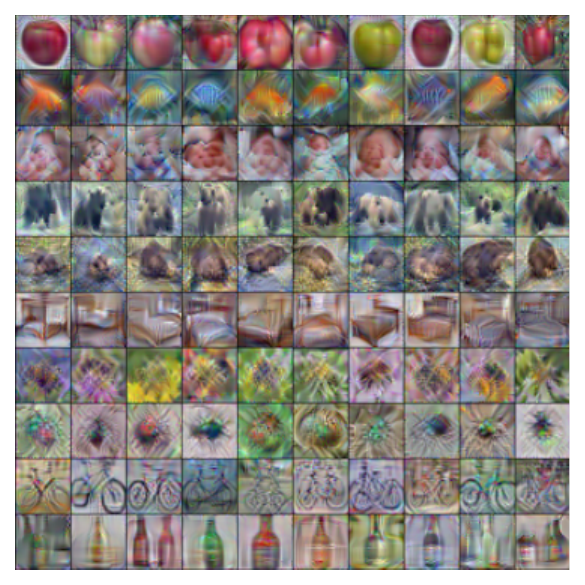}
        \caption{The synthetic images of ours.}
        \label{fig:ours}
    \end{subfigure}
    \caption{Visualizations of original images, and synthetic images generated by MTT and our proposed methods.}
    \label{fig:img_visual}
\end{figure*}

\textbf{Models.}
We employ the ConvNet architecture in the distillation process, in line with other methods except for FrePo, which utilizes a different model that doubles the number of filters when the feature map size is halved. Our architecture comprises 128 filters in the convolutional layer with a 3 × 3 kernel, followed by instance normalization, ReLU activation, and an average pooling layer with a 2 × 2 kernel and stride 2. For CIFAR-10 and CIFAR-100, we adopt 3-layer convolutional networks (ConvNet-3). In the case of the Tiny ImageNet dataset with a resolution of 64 × 64, we employ a depth-4 ConvNet. For the ImageNet subsets with a resolution of 128 × 128, a depth-5 ConvNet is used.

\textbf{Implementation Details.} 
In the buffer phase, we utilize SGD with momentum as the optimizer, setting $\lambda$ as an array ranging from 0.5 to 1. This range is applied individually for the first 5 training epochs, while $\mu$ is set to 1. We train each expert model for 50 epochs, with the learning rate reduced by half in the 25th epoch. During the distillation process, the value of K in the K-Means algorithm is determined based on the ipc (images per class) value. However, when ipc equals 50, we opt to cluster only 10 sub-clusters, each selecting extra 4 points approximating the sub-cluster centroid. For the balanced loss, we set $\theta$ to 8. Concerning the intermediate matching loss, we introduce a hyperparameter $\beta$ to control the scale of several losses, encompassing two strategies: equal scale $\beta$=1 or varied scale $\beta$ based on the value of $\{\xi\}$. In terms of weight perturbation, we set $\rho$ to 0.1, and we also conduct experiments to evaluate the performance with dropout added. Throughout the evaluation phase, the number of training iterations is set to 1000, with a learning rate reduction by half at the 500-th iteration. Furthermore, we employ the same Differentiable Siamese Augmentation (DSA) during both the distillation and evaluation processes. 

\subsection{Comparison with State-of-the-Art Methods}

\textbf{Competitors}: We categorize the current works into three main parts, shown in \cref{tab:main-result}. The first part focuses on coreset selection and includes non-learning methods such as random selection, herding methods \cite{chen2012super}, and example forgetting \cite{Forgetting}. These techniques aim to select ``valuable'' instances from a large dataset. The second part comprises methods like DC, DM, DSA, CAFE, and FrePo, which primarily address short-range matching and truncated gradient backpropagation. They employ gradient matching or distribution matching as the optimization objective. Within the third part, MTT, TESLA, and FTD all utilize long-range matching characteristics based on the trajectory matching framework. However, MTT remains our primary competitor since TESLA and FTD integrate additional optimization techniques.

\textbf{Results with Coreset \& Short-range}: Our proposed method outperforms the coreset selection baselines significantly. The non-learning methods achieve inferior performance compared to our methods. When comparing with the second part works, our method demonstrates the same superiority over all other methods. Moreover, we improve one of the strong baseline DSA to nearly 15\% accuracy on both CIFAR-10 and CIFAR-100 under all ipc settings.

\textbf{Results with Long-range}: Regarding the comparison of experimental results in the third part, we observe consistently positive outcomes with significant improvements in all settings. We specifically compare our method with the original MTT. For example, in CIFAR-10, at a compression rate of fifty images, we achieve a score of 74.6\%, which is 3\% higher than the accuracy score of MTT. Compared to the results obtained from the full dataset, we are only around 10\% behind. Similar improvements are observed in CIFAR-100, where we achieve a 4\% increase over MTT and our results are only 4.5\% lower than the results obtained from the full dataset. We visualize parts of the distilled images in \cref{fig:img_visual} and discover that our distilled samples \cref{fig:ours} can focus more on the classified object itself while diluting the background information. 

\begin{figure}
    \centering
    \includegraphics[width=0.9\linewidth]{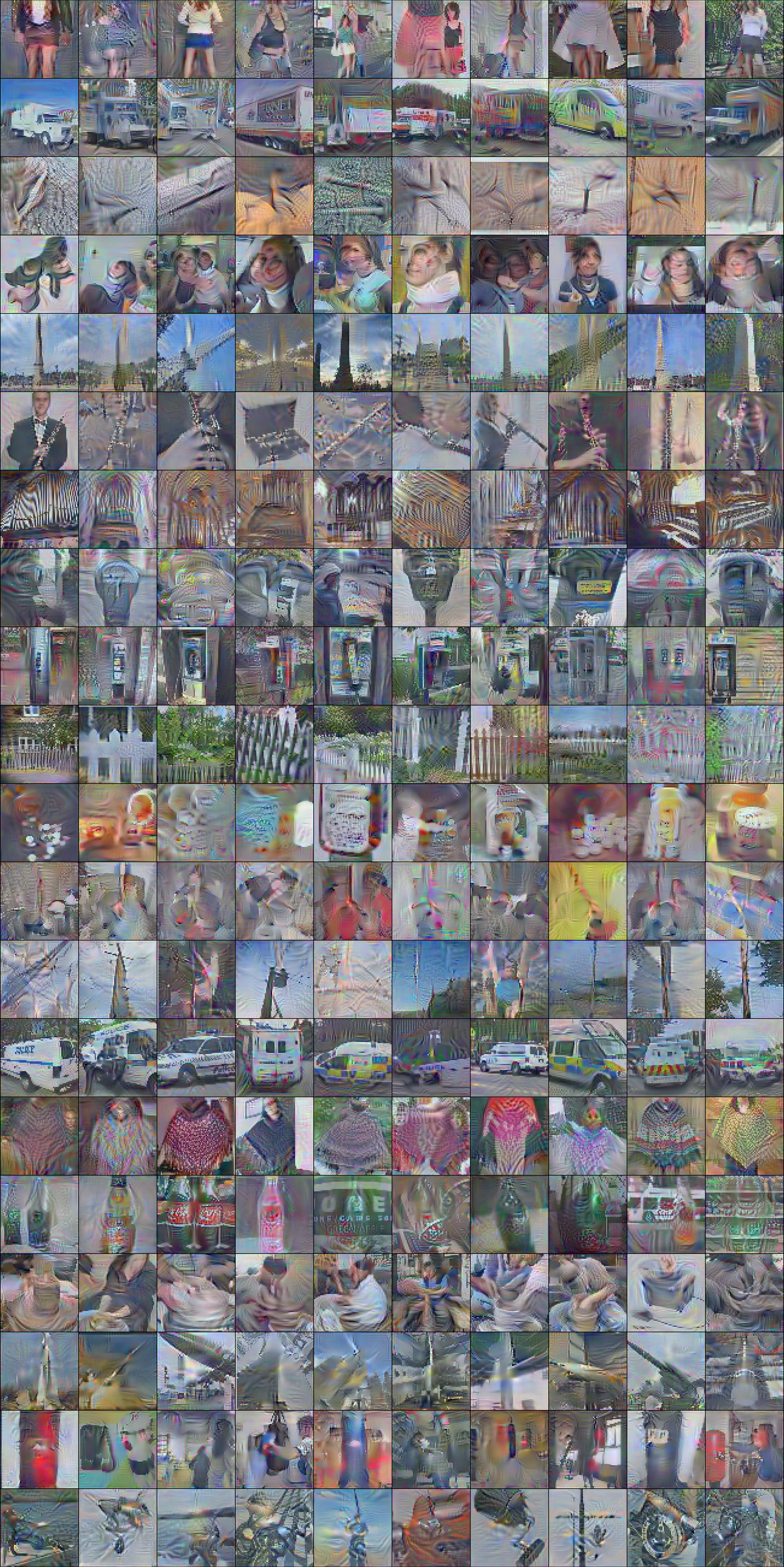}
    \caption{Visualizations of synthetic images in Tiny ImageNet.}
    \label{fig:tiny}
\end{figure}

\begin{table}[t]
\renewcommand\arraystretch{0.75}
\scriptsize
\begin{center}
\resizebox{0.95\linewidth}{!}{%
\begin{tabular}{@{}cc|cccc@{}}
\toprule  &&\multicolumn{4}{c}{Evaluation Model} \\
      && ConvNet           & ResNet           & VGG             & AlexNet          \\ \midrule
\multirow{5}{*}{\rotatebox{90}{Method}} &DC              & 53.9±0.5 &20.8±1.0 &38.8±1.1 &28.7±0.7         \\ 
&CAFE             & 55.5±0.4 &25.3±0.9 &40.5±0.8 &34.0±0.6 \\
&MTT             & 71.6±0.2 &61.9±0.7 &55.4±0.8 &48.2±1.0 \\ 
&FTD             & 73.8±0.2 &65.7±0.3 &58.4±1.6 &53.8±0.9 \\ 
&\textbf{Ours}             & \textbf{74.6±0.5} & \textbf{67.3±0.4} & \textbf{60.3±0.5} & \textbf{56.7±0.3} \\ \bottomrule
\end{tabular}%
}
\end{center}
\captionsetup{skip=-2pt} 
\caption{Generalization testing of different architectures on CIFAR-10 dataset with IPC 50.}
\label{tab:cross}
\end{table}

\begin{table}[t]
\centering
\Large
\renewcommand\arraystretch{1.42}
\resizebox{\columnwidth}{!}{%
\begin{tabular}{@{}c|cc|cc|cc|cc@{}}
\toprule
     \multirow{2}{*}{IPC}        & \multicolumn{2}{c|}{\textbf{ImageNette}}   & \multicolumn{2}{c|}{\textbf{ImageWoof}}     & \multicolumn{2}{c|}{\textbf{ImageFruit}}   & \multicolumn{2}{c}{\textbf{ImageMeow}} \\ 
          & 1        & 10           & 1       & 10              & 1   & 10        & 1              & 10               \\ \midrule
Full dataset & \multicolumn{2}{c|}{87.4±1.0}   & \multicolumn{2}{c|}{67.0±1.3}      & \multicolumn{2}{c|}{63.9±2.0}   & \multicolumn{2}{c}{66.7±1.1}     \\ \midrule
MTT~\cite{mtt}       & 47.7±0.9 & 63.0±1.3 & 28.6±0.8 & 35.8±1.8 & 26.6±0.8     & 40.3±1.3 & 30.7±1.6          & 40.4±2.2          \\
FTD~\cite{du2022minimizing}& 52.2±1.0 & 67.7±0.7 & 30.1±1.0 & 38.8±1.4 & 29.1±0.9     & 44.9±1.5 & 33.8±1.5          & 43.3±0.6         \\ \midrule
\textbf{Ours}                  & \textbf{53.1±0.8}  & \textbf{68.4±1.2} & \textbf{31.6±0.6} & \textbf{39.5±1.5} & \textbf{30.0±1.2} & \textbf{45.4±1.5} & \textbf{34.6±1.5} & \textbf{44.9±1.7}             \\ \bottomrule
\end{tabular}%
}
\caption{Applying our methods to 128$\times$128 resolution ImageNet subsets. Bold digits represent the best results.}
\label{imagenet}
\end{table}

In the more intricate Tiny ImageNet dataset, our approach demonstrates consistent and significant improvements, achieving 13.7\% and 25.7\% in ipc values of one and ten. These empirical findings substantiate the efficacy of our proposed method. We visualize parts of the distilled images in \cref{fig:tiny}.

\textbf{Cross-Architecture Generalization}: We evaluate generalization capacity across various architectures. Initially, we distilled the synthetic dataset using ConvNet. Subsequently, we trained several architectures, namely AlexNet, VGG11, and ResNet18, on the distilled dataset. ~\cref{tab:cross} presents the results of our evaluations. Our method outperforms MTT significantly in generalization performance.




\begin{figure*}[t]
    \centering
    \includegraphics[width=0.85\linewidth]{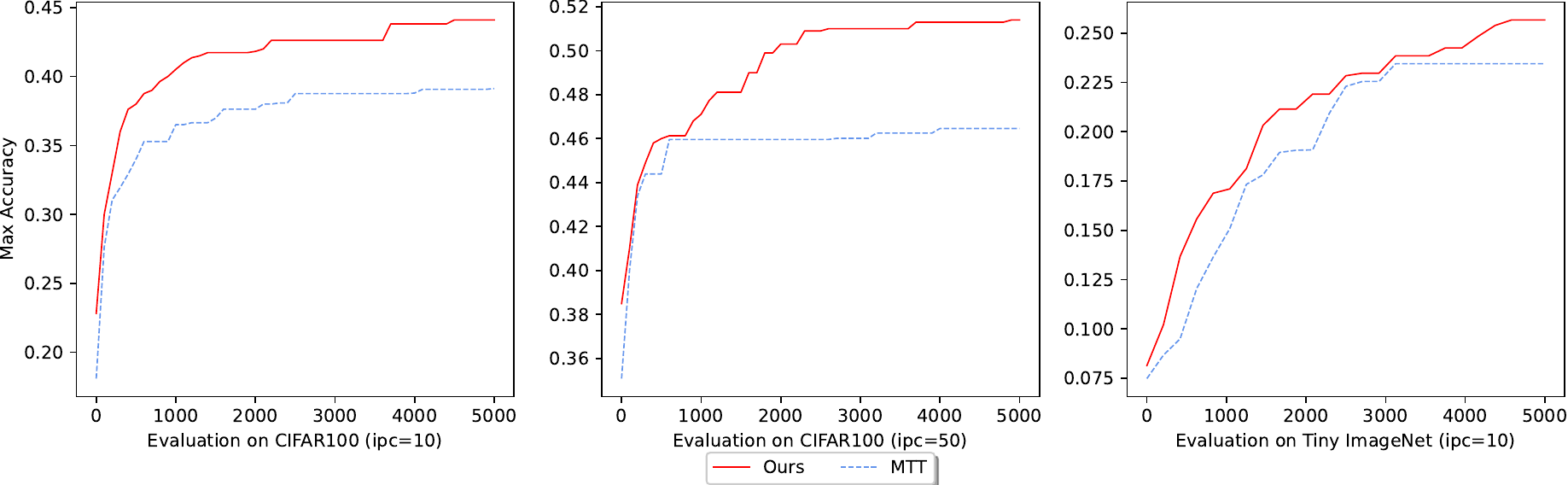}
    \caption{Applying our proposed methods brings stable performance and efficiency improvements.}
    \label{fig:acc_curves}
\end{figure*}

\begin{table}[b]
\centering
\small
\resizebox{1\linewidth}{!}{%
\begin{tabular}{@{}cc|c|cc@{}}
\toprule
Dataset                        & Image per class & 1 Iter. (sec) & 1k Iter. (min) & 5k Iter. (min)\\ \midrule
\multirow{3}{*}{CIFAR-10}      & 1   & 0.5  & 8.3 & 41                                                    \\
                               & 10  & 0.6  & 11  & 50                                                  \\
                               & 50  & 0.9 & 15 & 75                                                     \\ \midrule
\multirow{3}{*}{CIFAR-100}     & 1   & 0.6 & 11 & 50                                                    \\
                               & 10  & 0.85 & 14 & 70                                                     \\
                               & 50  & 1.97 & 33 & 163                                                     \\ \midrule
\multirow{2}{*}{Tiny ImageNet} & 1   & 1.15 & 20& 95                                                    \\
                               & 10  & 2.42  & 40 &200                                                   \\ \bottomrule
\end{tabular}%
}
\caption{Distillation time for each dataset and support IPC.}
\label{tab:timeburden}
\end{table}

\subsection{Results on ImageNet Subsets (128$\times$128)}
To further evaluate our method, we present results on a larger and more challenging dataset in \cref{imagenet}. The ImageNet subsets pose a greater difficulty compared to CIFAR-10/100 and Tiny ImageNet, primarily due to their higher resolutions. The higher resolution makes it challenging for the distillation procedure to converge. The ImageNet subsets consist of 10 categories selected from ImageNet-1k, following the setting of MTT. These subsets include ImageNette (assorted objects), ImageWoof (dog breeds), ImageFruits (fruits), and ImageMeow (cats). As shown in ~\cref{imagenet}, our method significantly improves MTT in every subset. For instance, we achieve a significant performance boost on the ImageNette subset with ipc = 1 and 10, surpassing MTT by more than 5.0\%. We also record the FTD results for fair comparison and our method achieves much better.

\subsection{Training Burden}

While we have introduced additional plug-in modules into the distillation process, the training time for each image in our approach remains comparable to that of the original MTT. It only requires a slight increase in time for each iteration, as illustrated in \cref{tab:timeburden}.

However, notably, our method exhibits faster convergence and superior results. As depicted in \cref{fig:acc_curves}, our approach on CIFAR100 essentially attains the final performance of the original MTT after just 500 iterations, while the original MTT requires 5000 iterations to achieve the same level of performance. Moreover, our method consistently demonstrates improvement. For example on TinyImageNet, our approach's performance continues to ascend, in contrast to MTT, which essentially plateaus after 3000 iterations.


\begin{figure}
    \centering
    \includegraphics[width=1\linewidth]{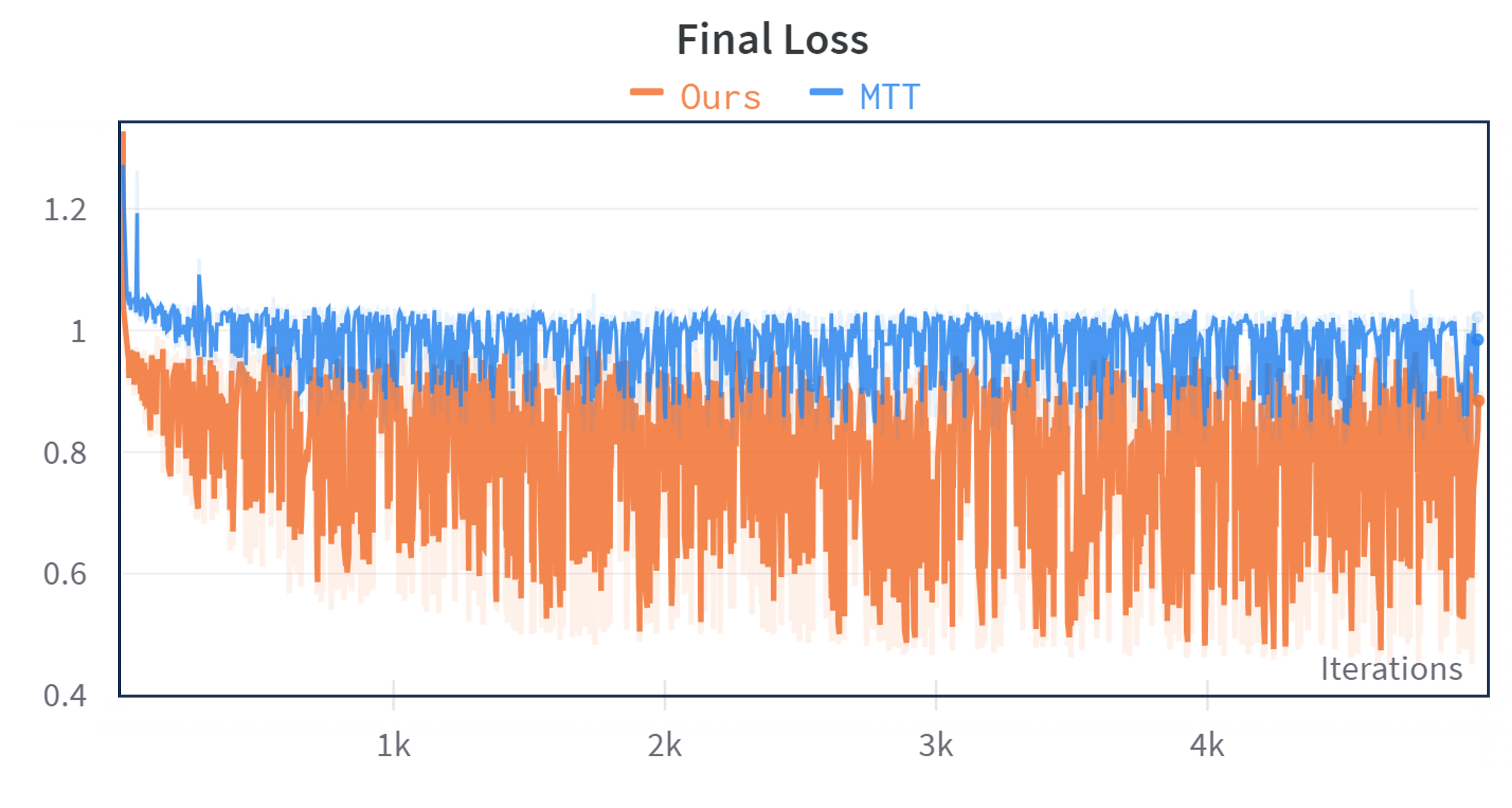}
    \caption{Comparison between proposed methods and original MTT. Our methods can generate a much lower final loss.}
    \label{fig:final_loss}
\end{figure}

\begin{figure}
    \centering
    \begin{subfigure}{0.49\linewidth}
        \centering
        \includegraphics[width=\linewidth]{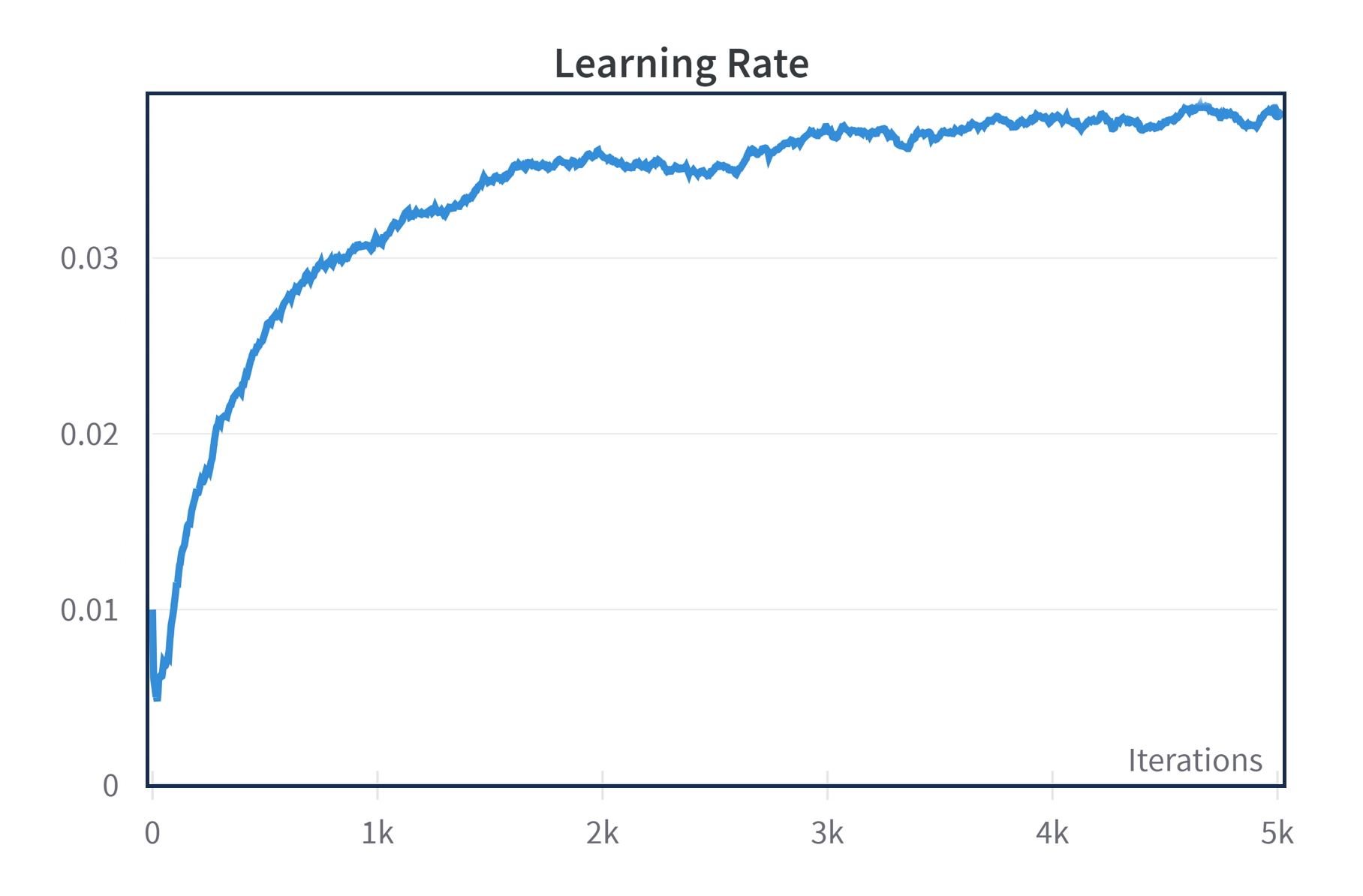}
        \caption{Smooth expert trajectory}
        \label{fig:lr_good}
    \end{subfigure}
    \begin{subfigure}{0.49\linewidth}
        \centering
        \includegraphics[width=\linewidth]{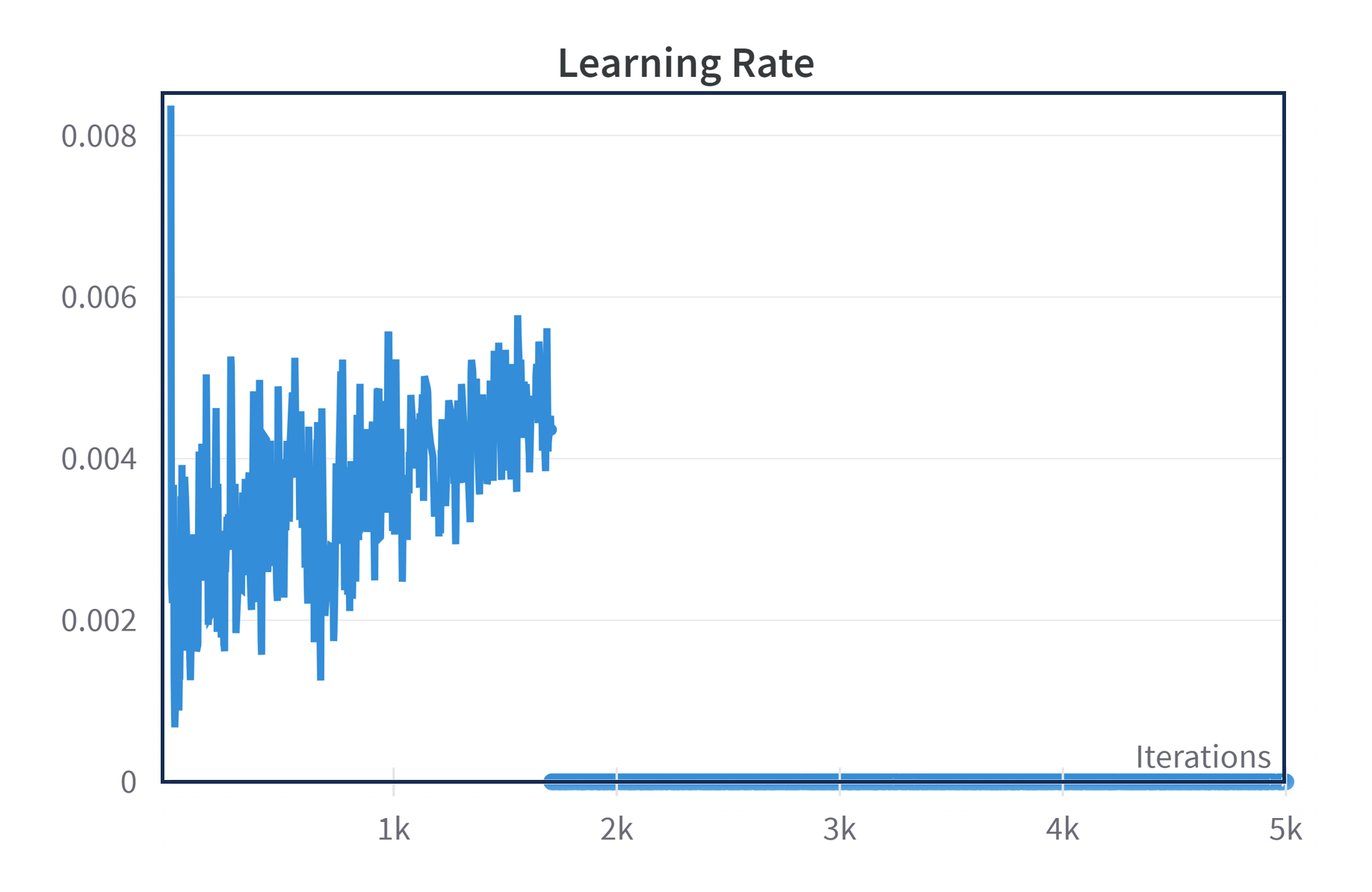}
        \caption{Non-smooth expert trajectory }
        \label{fig:lr_bad.}
    \end{subfigure}

    \caption{Learning curve for student model learning rate. When applying a non-smooth expert trajectory, the output of the learning rate may encounter NaN which will lead to the collapse of the training process.}
    \label{fig:lr}
\end{figure}


\begin{figure*}[]
    \centering
    \begin{subfigure}{0.48\linewidth}
        \centering
        \includegraphics[width=0.95\linewidth]{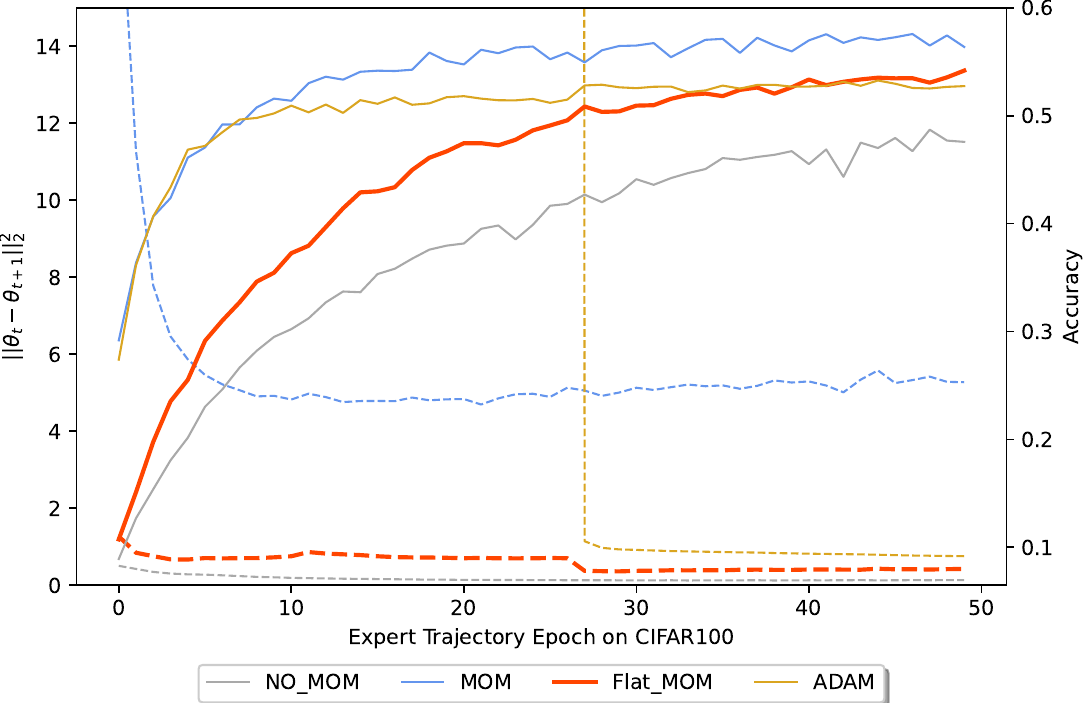}
        \caption{CIFAR-100}
        \label{fig:flat_show_cifar100}
    \end{subfigure}
    \begin{subfigure}{0.48\linewidth}
        \centering
        \includegraphics[width=0.95\linewidth]{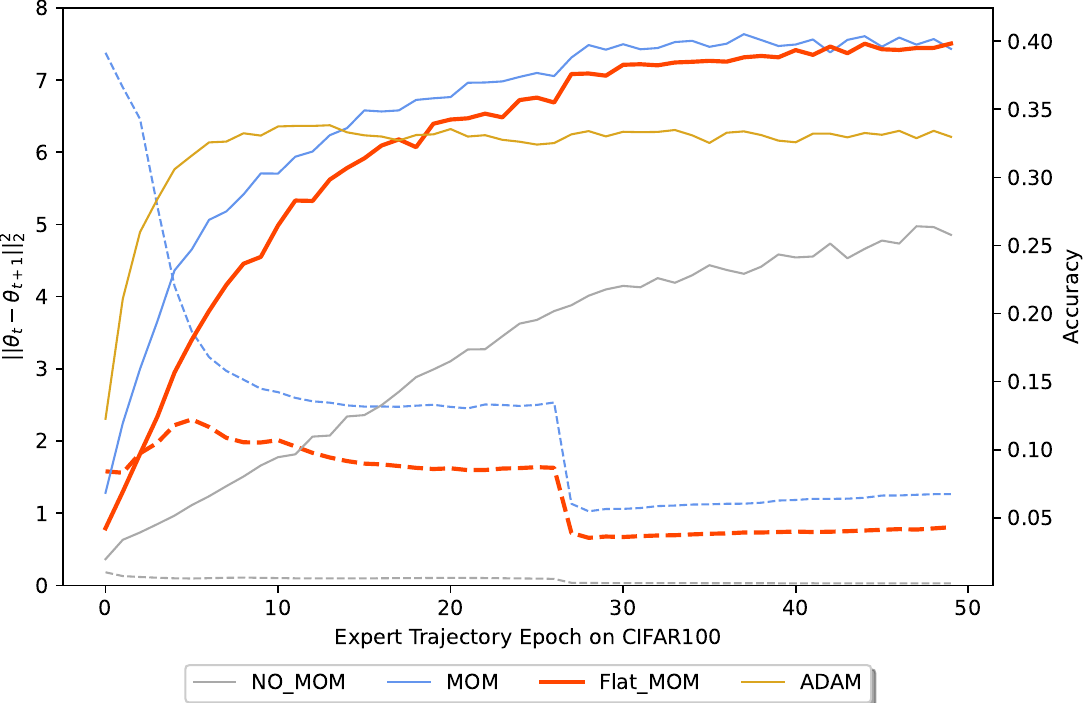}
        \caption{Tiny ImageNet}
        \label{fig:flat_show_tiny}
    \end{subfigure}
    \caption{${\left\|{\theta}_{t}-{\theta}_{t+1}\right\|_{2}^{2}}$ refers to the change of the model weights on two consecutive iterations, shown by dash curve. Correspondingly, the solid curve refers to the metric of evaluation accuracy. NO\_MOM refers to SGD without momentum. MOM means using SGD with momentum alone. Flat\_MOM denotes smooth expert trajectories that apply gradient penalty and clipped loss under the usage of SGD with momentum. ADAM means using ADAM as an optimizer alone. We view the red curve as a much smoother and higher-quality expert trajectory. The ${\left\|{\theta}_{t}-{\theta}_{t+1}\right\|_{2}^{2}}$ of Adam is so huge that it cannot fully appear in both CIFAR-100 and Tiny ImageNet.}
    \label{fig:flat_show}
\end{figure*}

\begin{table*}[]
    \begin{subtable}{.49\linewidth}
    \begin{center}
        \resizebox{\linewidth}{!}{%
        \begin{tabular}{@{}cccccc@{}}
        \toprule 
        \textbf{Momentum} & \begin{tabular}[c]{@{}c@{}}\textbf{Gradient}\\ \textbf{Penalty}\end{tabular}&\begin{tabular}[c]{@{}c@{}}\textbf{Clipped}\\ \textbf{Loss}\end{tabular}& \textbf{Avg\_Var} $\downarrow$ & \textbf{Acc. (Expert)} $\uparrow$ & \textbf{Acc. (Distill)} $\uparrow$ \\ \midrule
         $\times$ & $\times$ & $\times$          & 0.1726            & 48.6             & 39.7                    \\
        \checkmark & $\times$ & $\times$            & 7.9611 ($\times$46)            & 57.1 ($+$8.5)            & 18.8 ($-20.9$)                   \\
        \checkmark & \checkmark & $\times$      & 0.9331 ($\times$5.4)           & 54.1 ($+$5.5)            & 41.7 ($+$2.0)                   \\
        \checkmark & \checkmark & \checkmark        & \textbf{0.5899} ($\times$3.4)   & \textbf{54.4} ($+$5.8)    & \textbf{42.0} ($+$2.3)          \\ \bottomrule
        \end{tabular}
        }
    \end{center}
    \caption{CIFAR-100}
    \label{tab:cifar100var}
    \end{subtable}
    \hspace{6pt}
    \begin{subtable}{.49\linewidth}
    \begin{center}
        \resizebox{\linewidth}{!}{%
        \begin{tabular}{@{}cccccc@{}}
        \toprule 
        \textbf{Momentum} & \begin{tabular}[c]{@{}c@{}}\textbf{Gradient}\\ \textbf{Penalty}\end{tabular}&\begin{tabular}[c]{@{}c@{}}\textbf{Clipped}\\ \textbf{Loss}\end{tabular}& \textbf{Avg\_Var} $\downarrow$ & \textbf{Acc. (Expert)} $\uparrow$ & \textbf{Acc. (Distill)} $\uparrow$ \\ \midrule
         $\times$ & $\times$ & $\times$          & 0.0705            &  25.8            &  8.8                   \\
        \checkmark & $\times$ & $\times$            & 2.2950 ($\times$33)            & 39.4 ($+$13.6)            & 1.8 ($-7.0$)                   \\
        \checkmark & \checkmark & $\times$      &   1.7358 ($\times$24)           & 39.0 ($+$13.2)            & 10.0 ($+$1.2)                   \\
        \checkmark & \checkmark & \checkmark        & \textbf{1.3066} ($\times$18)   & \textbf{39.5} ($+$13.7)    & \textbf{10.8} ($+$2.0)          \\ \bottomrule
        \end{tabular}%
        }
    \end{center}
    \caption{Tiny ImageNet}
    \label{tab:tinyvar}
    \end{subtable} 
    \caption{Comparison between different buffer generation methods using SGD as base optimizer.} 
    \label{tab:avg_var}
\end{table*}

\subsection{Examples of Training Instability}
The sources of training instability can be attributed to two main factors. Firstly, the original MTT itself is prone to encountering sudden spikes in gradients, which can lead to training collapse. However, by incorporating the proposed methods, our training process becomes more stable. As illustrated in \cref{fig:final_loss}, our final loss is substantially lower than that of the original MTT, indicating a more effective transfer of expert network knowledge into the target compressed dataset.

The second source of instability stems from the parameter variations in the expert model trajectories. We utilize simple momentum-based SGD as the optimizer for training the expert model. Subsequently, we compare a crucial output during each iteration: the learnable variable ``learning rate''. From \cref{fig:lr}, it becomes evident that as the expert trajectories become less smooth, the learnable quantity experiences huge fluctuations. At around 1800 iterations, it even reaches NaN values due to sudden gradient explosions. As the previous theoretical analysis \cref{sec:whypoor} indicated, the occurrence of alignment failure is due to the continuous amplification of the accumulated gradient by the momentum. This also emphasizes the importance of generating smooth, slowly changing, and high-quality expert trajectories.

\subsection{Analysis}
\subsubsection{The Impact of Expert Trajectory Smoothness}
\label{buffer_ana}

\cref{fig:flat_show} and \cref{tab:avg_var} elucidate why previous works have invariably opted for naive SGD as the optimizer. This choice of SGD strikes an unavoidable trade-off between the performance of expert model and outcome of the distilled dataset. For instance, as shown in the ~\cref{tab:cifar100var}, the expert model alone achieves a modest score of 48.6\%, whereas distillation yields a respectable 39.7\%. However, despite adopting SGD with momentum or Adam enhancing expert performance (shown in blue and yellow solid lines), it leads to the variation of parameters changing so fast (shown in blue and yellow dash lines) that surpasses the limitation for distillation. Finally, it will cause significant declines in distillation results, even gradient explosions and training collapses, especially for Adam.

The essence of our proposed method for generating smooth expert trajectories lies in constraining the speed of parameter variation. The ideal expert trajectory exhibits slow parameter variations with consistent performance improvements along the iterations. To quantify smoothness, we employ a metric called $Avg\_Var$ to measure the average weight variation magnitude between two iterations along the whole training process:
\vspace{-0.1cm}
\begin{equation}
    Avg\_Var = {\mathbb{E}_t\left[\left\|{\theta}_{t}-{\theta}_{t+1}\right\|_{2}^{2}\right]} \nonumber
\end{equation}

\cref{fig:flat_show} shows ${\left\|{\theta}_{t}-{\theta}_{t+1}\right\|_{2}^{2}}$ of each epoch and \cref{tab:avg_var} demonstrates the average result. Through employing gradient penalty and loss clipping, we achieve significant improvements in both expert performance (an increase of 5.8\% and 13.7\%) and distilled dataset performance (an increase of 2.3\% and 2.0\%) while only increasing $Avg\_Var$ by a factor of 3.4 and 18 (compared to 46$\times$ and 33$\times$ for direct momentum addition, which is just a small increment).

\begin{figure*}[]
    \centering
    \includegraphics[width=0.95\linewidth]{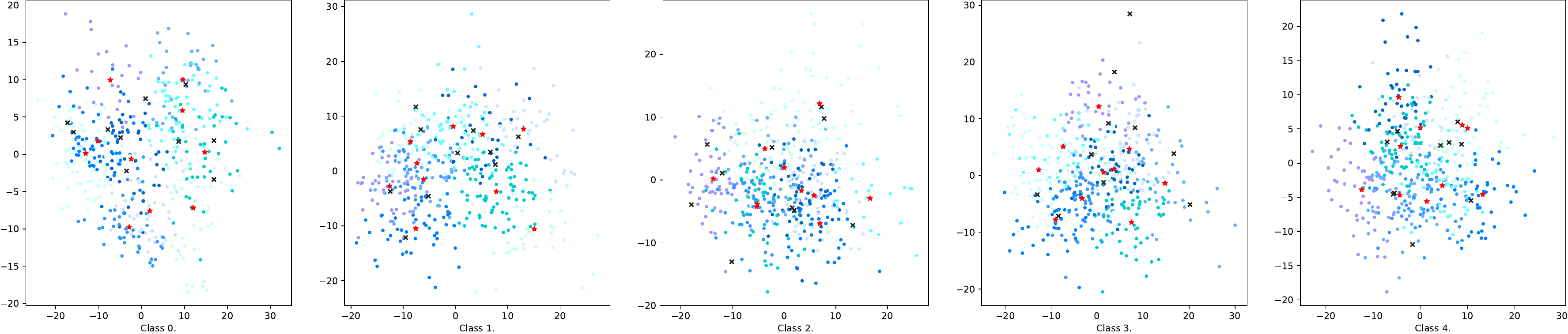}
    \caption{Example clustering and sub-cluster centre results. \textcolor{red}{$\star$} denotes the representative initialization samples while \textbf{$\times$} means the random initialization samples.}
    \label{fig:rep_show}
\end{figure*}

\begin{table}[]
\centering
\small
\resizebox{0.8\linewidth}{!}{%
\begin{tabular}{@{}lc@{}}
\toprule
\multicolumn{1}{c}{\textbf{Methods}} & \textbf{Acc. Distill (Gain)} \\ \midrule
Base                                 & 42.0                         \\ \midrule
Balance Stochasticity:               &                              \\
\quad+ Representative Initialization             & 42.5 (+0.5)                  \\
\quad+ Balanced Inner-loop Loss                      & 42.9 (+0.4)                  \\ \midrule
Alleviate Errors:                    &                              \\
\quad+ Intermediate Matching Loss                  & 43.6 (+0.7)                  \\
\quad+ Weight Perturbation                & \textbf{44.4 (+0.8)}         \\ \bottomrule
\end{tabular}%
}
\caption{We use the distilled results obtained by applying smooth expert trajectory as the ``Base''. Following that, we conduct two parts ablation studies (stochasticity and errors) on the CIFAR-100 dataset under IPC=10.}
\label{tab:abalation_last_two}
\end{table}

\begin{figure}[]
    \centering
    \includegraphics[width=0.95\linewidth]{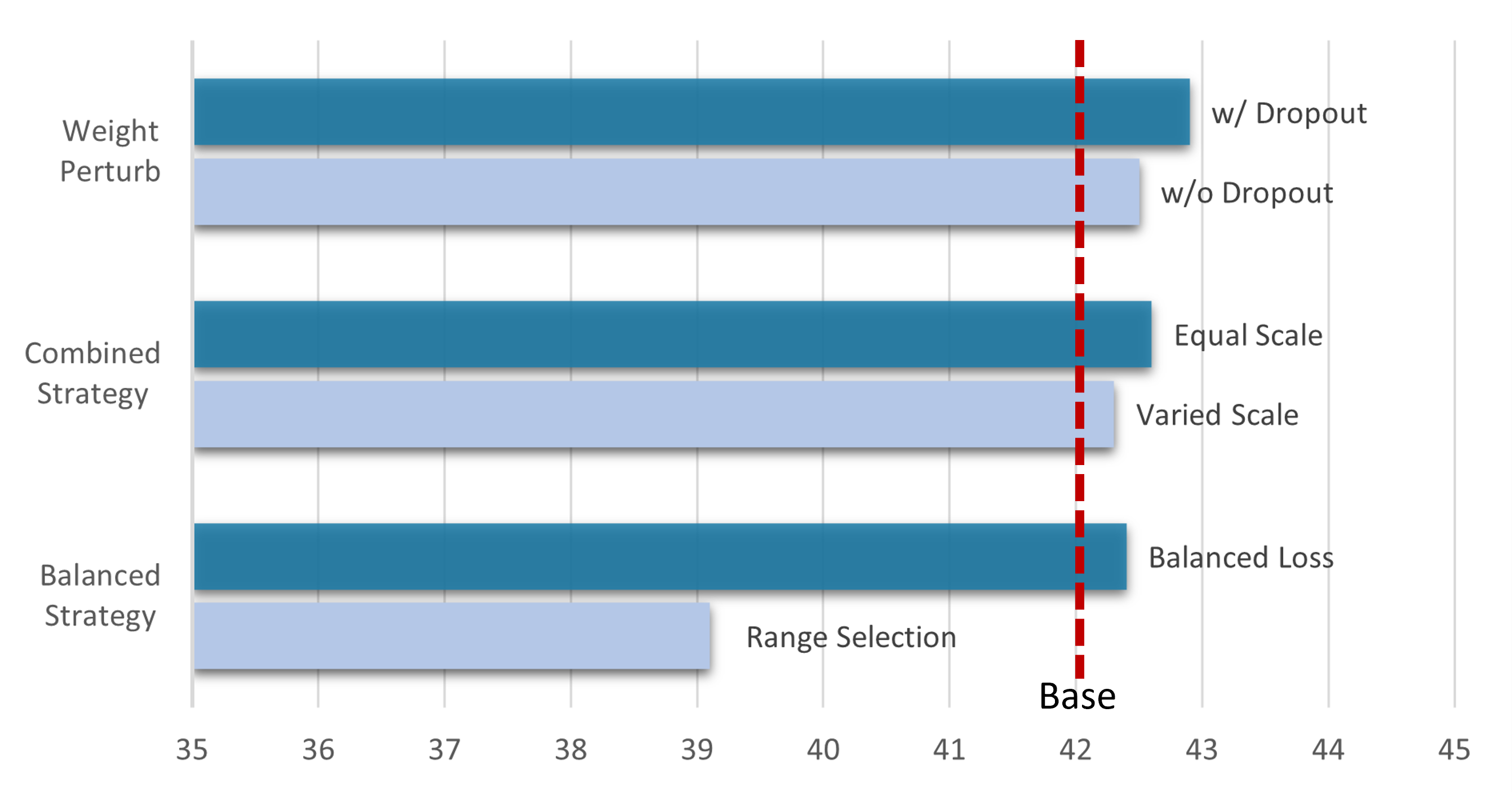}
    \caption{Results of ablation study on weight perturbation, intermediate matching loss and balanced strategy.}
    \label{fig:bar_abalation}
\end{figure}

\subsubsection{The Impact of Balance Strategy}
\label{Impact of Balance Strategy}
As demonstrated in \cref{tab:abalation_last_two}, we conduct ablation experiments on each proposed module based on the usage of smooth expert trajectory. In the ``Balance Stochasticity'' part, the act of selecting representative samples for initialization resulted in an incremental gain of 0.5\%. In order to visualize the effectiveness of the selection strategy, We randomly choose five classes and apply PCA~\cite{wold1987principal} to reduce the high-dimensional features to two dimensions. As we can see from \cref{fig:rep_show}, compared to random initialization, our proposed method avoids introducing huge bias coming from outliers. Additionally, the distribution of distilled samples is more uniform, and there is no over-concentration in a specific area.

Moreover, the application of a balanced inner-loop loss leads to a further enhancement, yielding an improvement of 0.4\%. As mentioned in \cref{balanceloss}, we compared another method: random initialization of $t$ within a certain range. We set the range to be within $±$5 when selecting the next expert starting point. From \cref{fig:bar_abalation}, the experiments indicate that initialization within a range not only lacks a positive effect but, conversely, introduces a potential erroneous inductive bias that results in a decline in distillation outcomes.

\subsubsection{The Impact of Accumulated Error}

To explore the contributions of the two error reduction methods, we conduct ablation experiments as depicted in \cref{tab:abalation_last_two}. In the ``Alleviate Errors'' part, the experiments indicate that using weight perturbation leads to the most improvement ($+$0.8\%) in distilled dataset performance. However, when applying $Dropout$ after weight perturbation shown in \cref{fig:bar_abalation}, which introduces another type of drop noise, we have not observed any improvement in the results.

Besides, intermediate matching loss contributes to an increase of 0.7\% to the final outcome, which elucidates that extended-range interactions indeed may lead to a divergence in the optimization direction within the inner loop. Reducing this part of the accumulated error is equally significant. We also experiment with two different strategies for combining multiple intermediate matching losses. As illustrated in \cref{fig:bar_abalation}, we observe that a straightforward approach, where the same scale of $\beta$ coefficient is used, yielded superior results.

\section{Conclusion}
In this paper, we propose a novel dataset distillation strategy AST to address the challenges associated with the mutual influence between expert and student models, sensitivity to stochastic variables, and accumulated errors. Building upon this, we argue the significant effect of trajectory smoothness and propose clipping loss and gradient penalty to smooth the expert trajectories under a more potent optimizer. The improved trajectories pave the way for optimizing the parameter alignment process from two perspectives. To temper the influence of two stochastic variables, we propose using representative initial samples for $\mathcal{D}_{syn}$ and replacing normal cross-entropy loss for balanced inner-loop loss. Besides, we propose intermediate matching loss and weight perturbation strategies to alleviate errors stemming from long-range inner steps $\mathcal{N}$ and discrepancies between distillation and evaluation.
Furthermore, our methods are designed to be easily implemented and plugged in, which broadens the scope of applicability. We hope AST can pave the way for future work on dataset distillation.

\bibliographystyle{IEEEtran}
\bibliography{mybibliography}

\end{document}